\documentclass[10pt,twocolumn,letterpaper]{article}

\usepackage{iccv}
\usepackage{times}
\usepackage{epsfig}
\usepackage{graphicx}
\usepackage{amsmath}
\usepackage{amssymb}

\usepackage[utf8]{inputenc} 
\usepackage[T1]{fontenc}    
\usepackage{url}            
\usepackage{booktabs}       
\usepackage{nicefrac}       
\usepackage{microtype}      
\usepackage{amsmath}
\usepackage[dvipsnames]{xcolor}
\usepackage{mathtools}

\usepackage{courier}
\usepackage{color, colortbl}
\usepackage{multirow}
\usepackage{graphicx} 
\usepackage{amsfonts}
\usepackage{bbm}
\usepackage{pifont}

\usepackage[font=small,labelfont=bf]{caption}
\usepackage[caption=false]{subfig}
\usepackage{wrapfig}
\usepackage[symbol]{footmisc}

\usepackage[title]{appendix}

\let\norm\undefined 
\DeclarePairedDelimiter\norm{\lVert}{\rVert}

\definecolor{mygray}{gray}{0.93}

\usepackage[pagebackref=true,breaklinks=true,letterpaper=true,colorlinks,bookmarks=false]{hyperref}

\iccvfinalcopy

\def\iccvPaperID{5861}

\begin{document}

\title{Hypercorrelation Squeeze for Few-Shot Segmentation}

\author{Juhong Min\hspace{0.9cm} Dahyun Kang\hspace{0.9cm} Minsu Cho\vspace{1.5mm}\\
Pohang University of Science and Technology (POSTECH), South Korea\\
{\tt\small \url{http://cvlab.postech.ac.kr/research/HSNet/}}
}

\maketitle


\begin{abstract}
Few-shot semantic segmentation aims at learning to segment a target object from a query image using only a few annotated support images of the target class. 
This challenging task requires to understand diverse levels of visual cues and analyze fine-grained correspondence relations between the query and the support images.
To address the problem, we propose Hypercorrelation Squeeze Networks (HSNet) that leverages multi-level feature correlation and efficient 4D convolutions.
It extracts diverse features from different levels of intermediate convolutional layers and constructs a collection of 4D correlation tensors, \ie, hypercorrelations.
Using efficient center-pivot 4D convolutions in a pyramidal architecture, the method gradually squeezes high-level semantic and low-level geometric cues of the hypercorrelation into precise segmentation masks in coarse-to-fine manner.
The significant performance improvements on standard few-shot segmentation benchmarks of PASCAL-5$^{i}$, COCO-20$^{i}$, and FSS-1000 verify the efficacy of the proposed method.
\end{abstract}
\vspace{-4.0mm}


\section{Introduction}
The advent of deep convolutional neural networks~\cite{he2016deep,huang2015dense,simonyan2015vgg} has promoted dramatic advances in many computer vision tasks including object tracking~\cite{li2019siammask,li2019simrpnpp,nam2015mdnet}, visual correspondence~\cite{huang2019dynamic,min2020dhpf,novotny2017anchornet}, and semantic segmentation~\cite{chen2018deeplabv3,noh2015learning,shelhamer2017fcn} to name a few.
Despite the effectiveness of deep networks, their demand for a heavy amount of annotated examples from large-scale datasets~\cite{deng2009imagenet, everingham2015pascal,lin2015coco} still remains a fundamental limitation since data labeling requires substantial human efforts, especially for dense prediction tasks, \eg, semantic segmentation.
To cope with the challenge, there have been various attempts in semi- and weakly-supervised segmentation approaches~\cite{chang2020weakly,koh2020sideinfnet,luo2020semiseg,sun2020mining,veksler2020regularized,wang2020selfweak,zhang2020splitting} which in turn effectively alleviated the data-hunger issue.
However, given only a {\em few} annotated training examples, the problem of poor generalization ability of the deep networks is yet the primary concern that many few-shot segmentation methods~\cite{dong2018pl, fan2020fgn, gairola2020simpropnet, hu2019amcg, li2020fss1000, liu2020crnet, liu2020ppnet, nguyen2019fwb, rakelly2018cofcn, shaban2017oslsm, siam2019amp, tian2020differentiable, tian2020pfenet, wang2020dan, wang2019panet, yang2020pmm, yang2020ltm, zhang2019pgnet, zhang2019canet, zhang2020sgone} struggle to address.

\begin{figure}[t]
    \centering
    \includegraphics[width=0.95\linewidth]{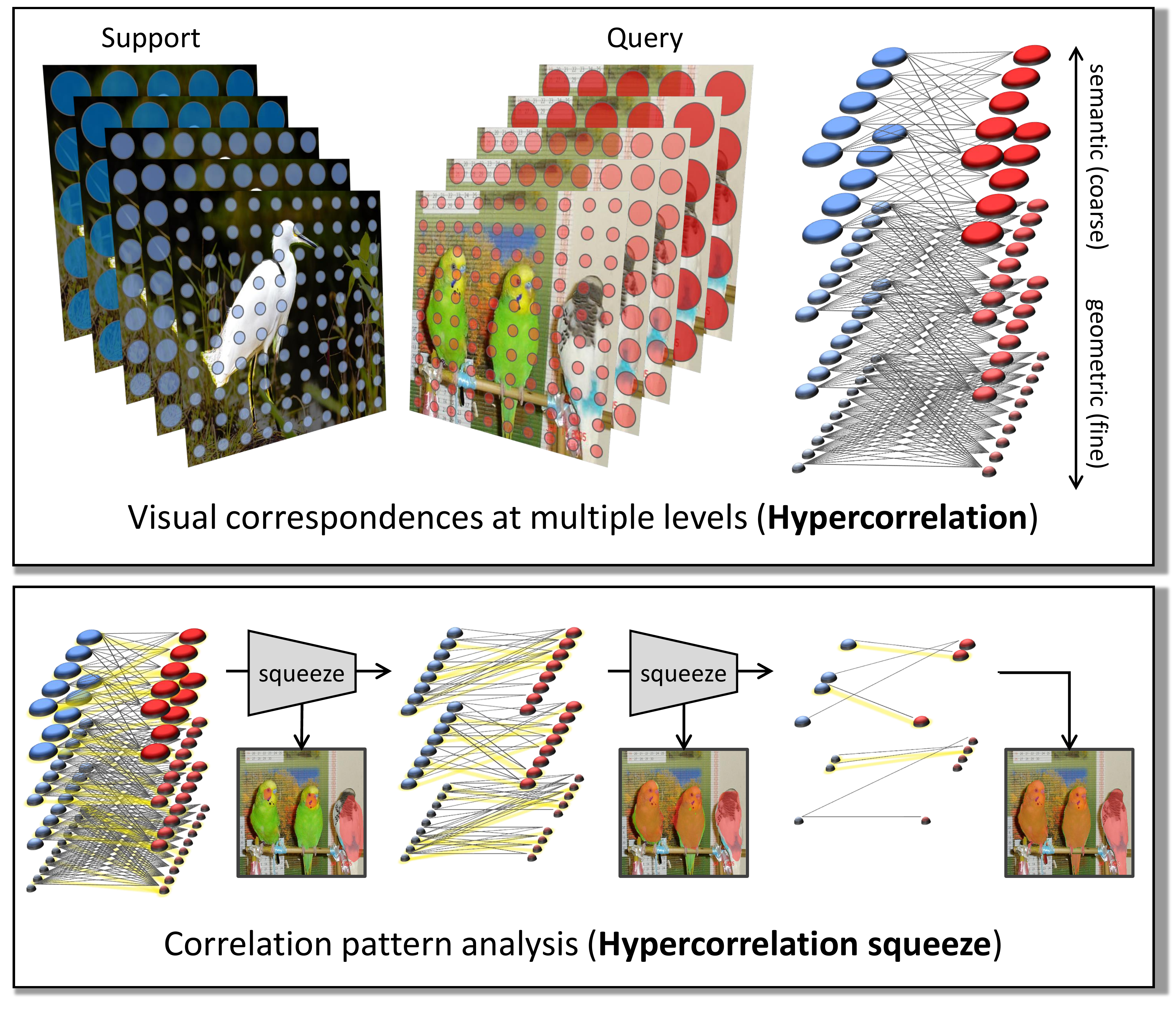}
	\vspace{-2.0mm}
	\caption{\label{fig:teaser}Our model performs visual reasoning in coarse-to-fine manner by gradually squeezing high-dimensional hypercorrelation to the target segmentation mask with efficient 4D convolutions.}
	\vspace{-7.0mm}
\end{figure}

In contrast, human visual system easily achieves generalizing appearances of new objects given extremely limited supervision.
The crux of such intelligence lies at the ability in finding reliable correspondences across different instances of the same class. Recent work on semantic correspondence shows that leveraging dense intermediate features~\cite{liu2020semantic,min2019hyperpixel,min2020dhpf} and processing correlation tensors with high-dimensional convolutions~\cite{li2020correspondence,rocco2018neighbourhood,truong2020glunet} are significantly effective in establishing accurate correspondences.
However, while recent few-shot segmentation research began active exploration in the direction of correlation learning, most of them~\cite{liu2020crnet,liu2020ppnet,nguyen2019fwb,jake2017prototypical,oriol2016oneshot,wang2019panet,yang2020pmm} neither exploit diverse levels of feature representations from early to late layers of a CNN nor construct pair-wise feature correlations to capture fine-grained correlation patterns. 
There have been some attempts~\cite{wang2020dan,zhang2019pgnet} in utilizing dense correlations with multi-level features, but they are yet limited in the sense that they simply employ the dense correlations for graph attention, using only a small fraction of intermediate conv layers.

In this work we combine the two of the most influential techniques in recent research of visual correspondence, multi-level features and 4D convolutions, and deign a novel framework, dubbed {\em Hypercorrelation Squeeze Networks} (HSNet), for the task of few-shot semantic segmentation.
As illustrated in Fig.~\ref{fig:teaser}, our network exploits diverse geometric/semantic feature representations from many different intermediate CNN layers to construct a collection of 4D correlation tensors, \ie, {\em hypercorrelations}, which represent a rich set of correspondences in multiple visual aspects.
Following the work of FPN~\cite{lin2017feature}, we adapt pyramidal design to capture both high-level semantic and low-level geometric cues for precise mask prediction in coarse-to-fine manner using deeply stacked 4D conv layers.
To reduce computational burden caused by such heavy use of high-dimensional convs, we devise an efficient 4D kernel via reasonable weight-sparsification which enables real-time inference while being more effective and light-weight than the existing ones.
The improvements on standard few-shot segmentation benchmarks of PASCAL-5$^{i}$~\cite{shaban2017oslsm}, COCO-20$^{i}$~\cite{lin2015coco}, and FSS-1000~\cite{li2020fss1000} verify the efficacy of the proposed method.

\section{Related Work}

\smallbreak
\noindent \textbf{Semantic segmentation.}
The goal of semantic segmentation is to classify each pixel of an image into one of the predefined object categories.
Prevalent segmentation approaches~\cite{cermelli2020modeling,chen2018deeplabv3,noh2015learning,olaf2015unet,peng2017global,shelhamer2017fcn,wang2020dualsuper} typically employ encoder-decoder structure in their architecture;
the encoder aggregates features along deep convolutional pathways and provides high-dimensional feature map in low-resolution and the corresponding decoder takes the output to predict segmentation mask by reversing this process~\cite{olaf2015unet}.
Although the methods clearly show the effectiveness of the encoder-decoder architecture in the task of semantic segmentation, offering useful insights to our study, they still suffer apparent disadvantages of data-driven nature of neural networks: lack of generalizibility under insufficient training data.

\smallbreak
\noindent \textbf{Few-shot learning.}
To resolve the generalization problem, many recent approaches to image classification made various attempts in training deep networks with a few annotated examples~\cite{allen2019infinite,hou2019cross,koch2015siamese,li2019revisiting,oreshkin2018tadam,qiao2019transductive,satorras2018few,jake2017prototypical,sung2018relationnet,oriol2016oneshot,wu2019parn,ye2020feat,zhang2020deepemd}.
Vinyals \etal~\cite{oriol2016oneshot} propose matching networks for one-shot learning; 
the method utilizes a special kind of mini-batches called episodes to match training and testing environments, facilitating better generalization on novel classes.
Snell \etal~\cite{jake2017prototypical} introduce prototypical networks which compute distances between representative embeddings, \ie, prototypes, for few-shot classification.
With the growing interests in few-shot learning in classification domain, the problem of few-shot segmentation has attracted a great deal of attention as well.
Shaban \etal~\cite{shaban2017oslsm} propose one-shot semantic segmentation networks which (meta-) learns to generate parameters of FCN~\cite{shelhamer2017fcn}.
Inspired by the prototypical networks~\cite{jake2017prototypical}, utilizing prototype representations to guide mask prediction in a query image became a popular paradigm in few-shot segmentation literature~\cite{dong2018pl,liu2020crnet,liu2020ppnet,nguyen2019fwb,siam2019amp,wang2019panet,yang2020pmm,zhang2019canet,zhang2020sgone}.
Witnessing the limitation of prototypical approaches, \eg, loss of spatial structure due to masked average pooling~\cite{zhang2020sgone}, work of~\cite{wang2020dan,zhang2019pgnet} build pair-wise feature correlations, \eg, graph attention, to retain the spatial structure of the images for fine-grained mask prediction.
Note that both prototypical and graph-based methods fundamentally focus on {\em learning to find reliable correspondences} between support and query images for accurate mask prediction.
In this work, we advance this idea and focus on {\em learning to analyze correspondences} using adequately designed learnable layers, \eg, 4D convolutions~\cite{rocco2018neighbourhood}, for effective semantic segmentation.

\smallbreak
\noindent \textbf{Learning visual correspondences.}
The task of visual correspondence aims to find reliable correspondences under challenging degree of variations~\cite{balntas2017hpatches,ham2016proposal,ham2018proposal, min2019spair,sattler2018benchmarking}.
Many methods~\cite{huang2020robust,huang2019dynamic,li2020correspondence,liu2020semantic,min2019hyperpixel,min2020dhpf,rocco17geocnn, rocco2018neighbourhood,yang2019optical} typically built upon convolutional features pretrained on classification task~\cite{deng2009imagenet}, showing they serve as good transferable representations.
Recent approaches to semantic correspondence~\cite{huang2020robust,liu2020semantic,min2019hyperpixel,min2020dhpf} show that efficiently exploiting different levels of convolutional features distributed over {\em all} intermediate layers clearly benefits matching accuracy.
In wide-baseline matching literature, a trending choice is to employ 4D convolutions~\cite{li2020correspondence, min2021chm, rocco2020sparsencnet, rocco2018neighbourhood, truong2020glunet} on dense feature matches to identify spatially consistent matches by analyzing local patterns in 4D space.
The use of multi-level features and relational pattern analysis using 4D convs are the two widely adopted techniques in the field of visual correspondence.

\begin{figure*}
    \begin{center}
        \includegraphics[width=1.0\linewidth]{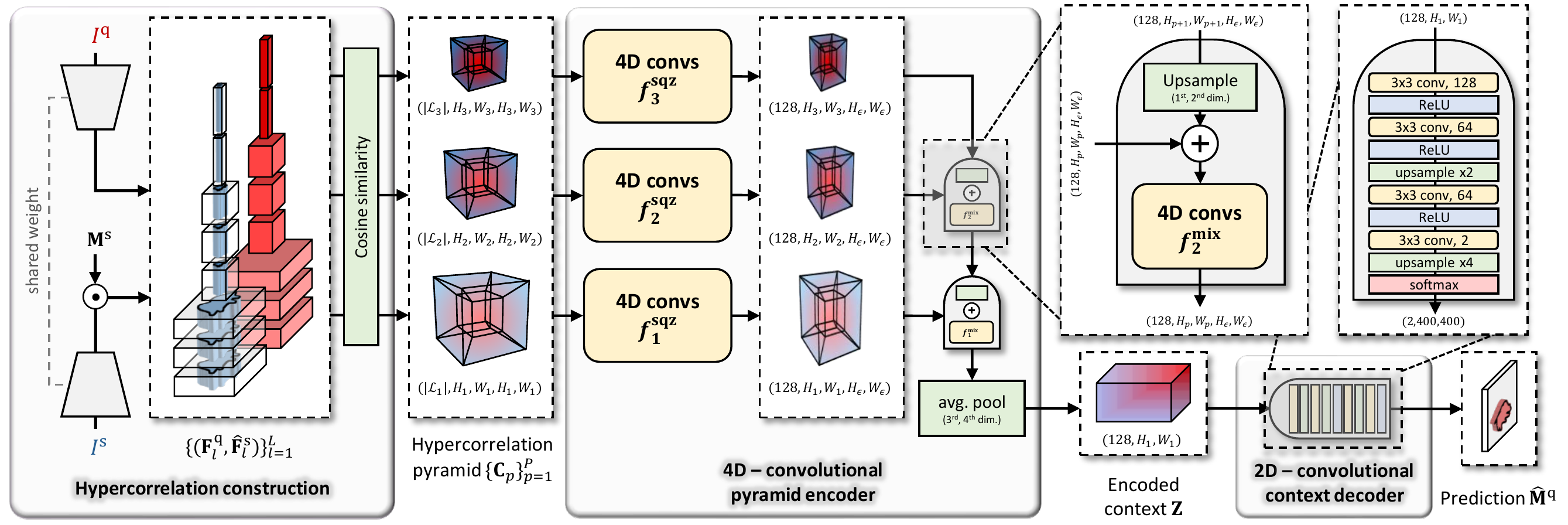}
    \end{center}
    \vspace{-5.0mm} 
      \caption{Overall architecture of the proposed network which consists of three main parts: hypercorrelation construction, 4D-convolutional pyramid encoder, and 2D-convolutional context decoder. We refer the readers to Sec.~\ref{sec:method} for details of the architecture.} 
    \vspace{-5.0mm} 
\label{fig:architecture}
\end{figure*}

In this paper we adapt the two most influential methodologies in visual correspondence to tackle few-shot segmentation: multi-level features and 4D convolutions. 
Inspired by the previous matching methods~\cite{min2019hyperpixel,min2020dhpf,lee2019sfnet}, which use multi-level features to build effective ``appearance features'',  we construct high-dimensional ``relational features'' using intermediate CNN features and process them with a series of 4D convolutions.
However, their quadratic complexity still remains a major bottleneck in designing cost-effective deep networks, constraining many previous matching methods~\cite{li2020correspondence, rocco2020sparsencnet, rocco2018neighbourhood, truong2020glunet} to use only a few 4D conv layers.
To resolve the issue, we develop a light-weight 4D convolutional kernel by collecting only a small subset of vital parameters for effective pattern recognition, which eventually leads to an efficient decomposition into a pair of 2D conv kernels with a linear complexity.
Our contributions can be summarized as follows:\vspace{-7px}
\begin{itemize}
\setlength\itemsep{-0.3em}
    \item We present the Hypercorrelation Squeeze Networks that analyze dense feature matches of diverse visual aspects using deeply stacked 4D conv layers.
    \item We propose center-pivot 4D conv kernel which is more effective than the existing one in terms both accuracy and speed, achieving real-time inference.
    \item The proposed method sets a new state of the art on three standard few-shot segmentation benchmarks: PASCAL-5$^{i}$~\cite{shaban2017oslsm}, COCO-20$^{i}$~\cite{lin2015coco}, and FSS-1000~\cite{li2020fss1000}.
\end{itemize}

\section{Problem Setup}
The goal of few-shot semantic segmentation is to perform segmentation given only a few annotated examples.
To avoid the risk of overfitting due to insufficient training data, we adopt widely used meta-learning approach called episodic training~\cite{oriol2016oneshot}.
Let us denote respective training and test sets as $\mathcal{D}_{\text{train}}$ and $\mathcal{D}_{\text{test}}$ which are disjoint with respect to object classes.
Both sets consist of multiple {\em episodes} each of which is composed of a support set $\mathcal{S} = ({I}^{\mathrm{s}}, \mathbf{M}^{\mathrm{s}})$ and a query set $\mathcal{Q} = ({I}^{\mathrm{q}}, \mathbf{M}^{\mathrm{q}})$ where ${I}^{*}$ and $\mathbf{M}^{*}$ are an image and its corresponding mask label respectively.
During training, our model iteratively samples an episode from $\mathcal{D}_{\text{train}}$ to learn a mapping from $({I}^{\mathrm{s}}, \mathbf{M}^{\mathrm{s}}, {I}^{\mathrm{q}})$ to query mask $\mathbf{M}^{\mathrm{q}}$.
Once the model is trained, it uses the learned mapping for evaluation without further optimization, \ie, the model takes randomly sampled $({I}^{\mathrm{s}}, \mathbf{M}^{\mathrm{s}}, {I}^{\mathrm{q}})$ from $\mathcal{D}_{\text{test}}$ to predict query mask. 


\section{Proposed Approach}
\label{sec:method}

In this section, we present a novel few-shot segmentation architecture, {\em Hypercorrelation Squeeze Networks} (HSNet), which capture relevant patterns in multi-level feature correlations between a pair of input images to predict fine-grained segmentation mask in a query image.
As illustrated in Fig.~\ref{fig:architecture}, we adopt an encoder-decoder structure in our architecture; 
the encoder gradually squeezes dimension of the input hypercorrelations by aggregating their local information to a global context, and the decoder processes the encoded context to predict a query mask.
In Sec.~\ref{sec:hpconstruction}-\ref{sec:decoder}, we demonstrate each pipeline in one-shot setting, \ie, the model predicts the query mask given ${I}^{\mathrm{q}}$ and $\mathcal{S} = ({I}^{\mathrm{s}}, \mathbf{M}^{\mathrm{s}})$. 
In Sec.~\ref{sec:s4d}, to mitigate large resource demands of 4D convs, we present a light-weight 4D kernel which greatly improves model efficiency in terms of both memory and time.
In Sec.~\ref{sec:extension_kshot}, we demonstrate how the model can be easily extended to $K$-shot setting, \ie, $\mathcal{S} = \{({I}^{\mathrm{s}}_{k}, \mathbf{M}^{\mathrm{s}}_{k})\}_{k=1}^{K}$, without loss of generality. 

\subsection{Hypercorrelation construction}
\label{sec:hpconstruction}

Inspired by recent semantic matching approaches~\cite{liu2020semantic,min2019hyperpixel,min2020dhpf}, our model exploits a rich set of features from the intermediate layers of a convolutional neural network to capture multi-level semantic and geometric patterns of similarities between the support and query images.
Given a pair of query and support images, ${I}^{\mathrm{q}}, {I}^{\mathrm{s}} \in \mathbb{R}^{3 \times H \times W}$, the backbone network produces a sequence of $L$ pairs of intermediate feature maps $\{(\mathbf{F}^{\mathrm{q}}_{l}, \mathbf{F}^{\mathrm{s}}_{l})\}_{l=1}^{L}$.
We mask each support feature map $\mathbf{F}^{\mathrm{s}}_{l} \in \mathbb{R}^{C_{l} \times H_{l} \times W_{l}}$ using the support mask $\mathbf{M}^{\mathrm{s}} \in \{0,1\}^{H \times W}$ to discard irrelevant activations for reliable mask prediction:
\begin{align}
\label{eqn:masking}
    \hat{\mathbf{F}}^{\mathrm{s}}_{l} = \mathbf{F}^{\mathrm{s}}_{l} \odot \zeta_{l}(\mathbf{M}^{\mathrm{s}}),
\end{align}
where $\odot$ is Hadamard product and $\zeta_{l}(\cdot)$ is a function that bilinearly interpolates input tensor to the spatial size of the feature map $\mathbf{F}^{\mathrm{s}}_{l}$ at layer $l$ followed by expansion along channel dimension such that $\zeta_{l}: \mathbb{R}^{H \times W} \xrightarrow{} \mathbb{R}^{C_{l} \times H_{l} \times W_{l}}$.
For the subsequent hypercorrleation construction, a pair of query and masked support features at each layer forms a 4D correlation tensor $\mathbf{\hat{C}}_{l} \in \mathbb{R}^{H_{l} \times W_{l} \times H_{l} \times W_{l}}$ using cosine similarity:
\begin{align}
\label{eqn:cosine}
    \mathbf{\hat{C}}_{l}(\mathbf{x}^{\mathrm{q}}, \mathbf{x}^{\mathrm{s}}) = \text{ReLU}\Bigg( \frac{\mathbf{F}^{\mathrm{q}}_{l}(\mathbf{x}^{\mathrm{q}}) \cdot \hat{\mathbf{F}}^{\mathrm{s}}_{l}(\mathbf{x}^{\mathrm{s}})} {\norm{\mathbf{F}^{\mathrm{q}}_{l}(\mathbf{x}^{\mathrm{q}})} \norm{\hat{\mathbf{F}}^{\mathrm{s}}_{l} (\mathbf{x}^{\mathrm{s}})}} \Bigg),
\end{align}
where $\mathbf{x}^{\mathrm{q}}$ and $\mathbf{x}^{\mathrm{s}}$ denote 2-dimensional spatial positions of feature maps $\mathbf{F}^{\mathrm{q}}_{l}$ and $\hat{\mathbf{F}}^{\mathrm{s}}_{l}$ respectively, and ReLU suppresses noisy correlation scores.
From the resultant set of 4D correlations $\{\mathbf{\hat{C}}_{l}\}_{l=1}^{L}$, we collect 4D tensors if they have the same spatial sizes and denote the subset as $\{\mathbf{\hat{C}}_{l}\}_{l \in \mathcal{L}_p}$ where $\mathcal{L}_{p}$ is a subset of CNN layer indices $\{1, ..., L\}$ at some pyramidal layer $p$.
Finally, all the 4D tensors in $\{\mathbf{\hat{C}}_{l}\}_{l \in \mathcal{L}_{p}}$ are concatenated along channel dimension to form a {\em hypercorrelation} $\mathbf{C}_{p} \in \mathbb{R}^{|\mathcal{L}_{p}| \times H_{p} \times W_{p} \times H_{p} \times W_{p}}$ where $(H_{p}, W_{p}, H_{p}, W_{p})$, with abuse of notation, represents the spatial resolution of the hypercorrelation at pyramidal layer $p$.
Given $P$ pyramidal layers, we denote {\em hypercorrelation pyramid} as $\mathcal{C}=\{\mathbf{C}_{p}\}_{p=1}^{P}$, representing a rich collection of feature correlations from multiple visual aspects.

\subsection{4D-convolutional pyramid encoder}
\label{sec:encoder}
\begin{figure}
    \begin{center}
        \includegraphics[width=0.95\linewidth]{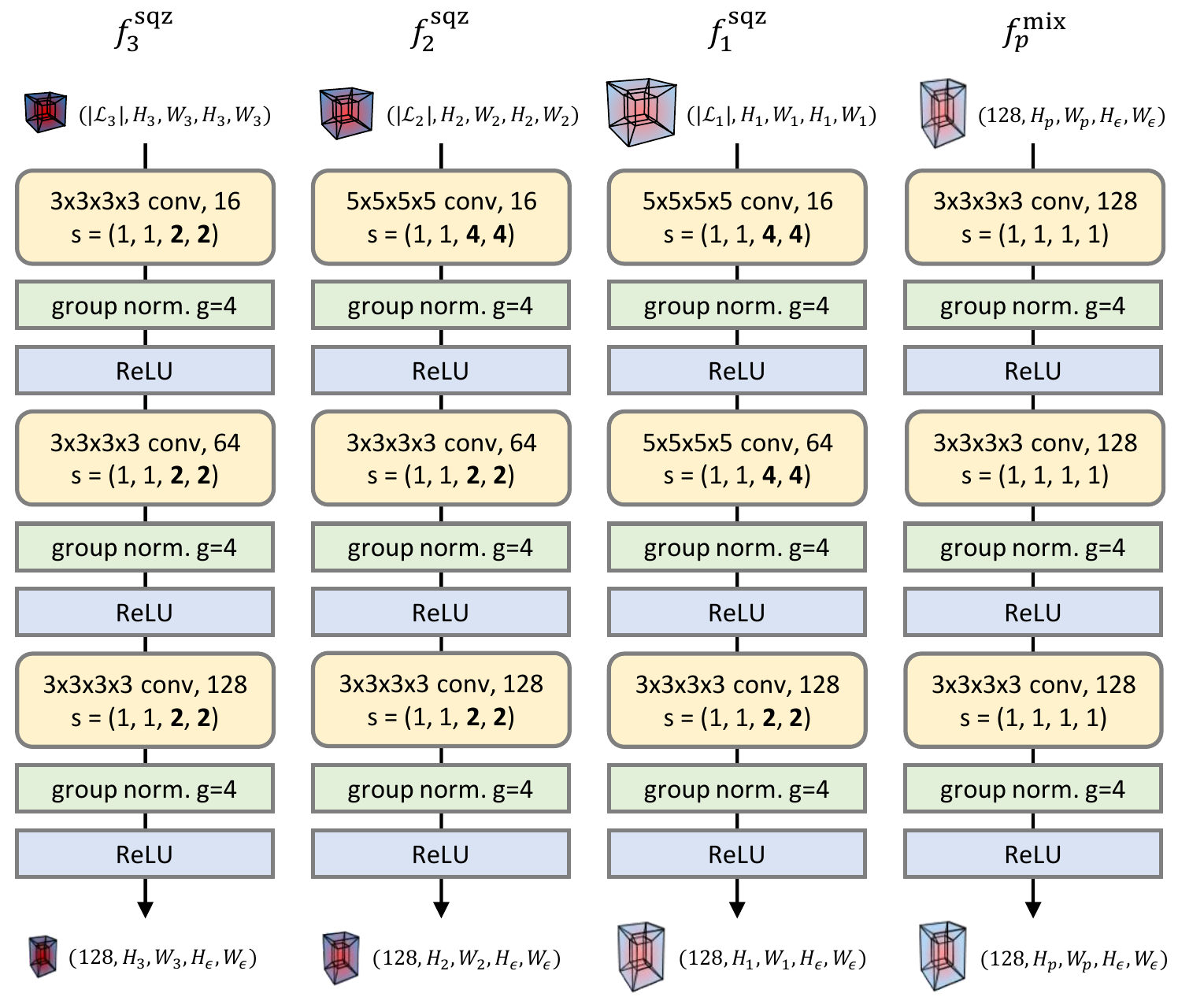}
    \end{center}
    \vspace{-5.0mm} 
      \caption{Building blocks in Hypercorrelation Squeeze Networks. $s$ and $g$ denotes strides of 4D conv and the number of groups in group normalization~\cite{wu2018groupnorm} respectively. Note $p \in \{1,2\}$ for $f_{p}^{\text{mix}}$.}
    \vspace{-3.0mm} 
\label{fig:cc4d}
\end{figure}

Our encoder network takes the hypercorrelation pyramid $\mathcal{C}=\{\mathbf{C}_{p}\}_{p=1}^{P}$ to effectively squeeze it into a condensed feature map $\mathbf{Z} \in \mathbb{R}^{128 \times H_{1} \times W_{1}}$.
We achieve this correlation learning using two types of building blocks: a squeezing block $f_{p}^{\text{sqz}}$ and a mixing block $f_{p}^{\text{mix}}$.
Each block consists of three sequences of multi-channel 4D convolution, group normalization~\cite{wu2018groupnorm}, and ReLU activation as illustrated in Fig.~\ref{fig:cc4d}.
In the squeezing block $f_{p}^{\text{sqz}}$, large strides periodically squeeze the last two (support) spatial dimensions of $\mathbf{C}_{p}$ down to $(H_{\epsilon}, W_{\epsilon})$ while the first two spatial (query) dimensions remain the same as $(H_{p}, W_{p})$, \ie, $f^{\text{sqz}}_{p}: \mathbb{R}^{|\mathcal{L}_{p}| \times H_{p} \times W_{p} \times H_{p} \times W_{p}} \xrightarrow{} \mathbb{R}^{128 \times H_{p} \times W_{p} \times H_{\epsilon} \times W_{\epsilon}}$ where $H_{p} > H_{\epsilon}$ and $W_{p} > W_{\epsilon}$.
Similar to FPN~\cite{lin2017feature} structure, two outputs from adjacent pyramidal layers, $p$ and $p+1$, are merged by element-wise addition after upsampling the (query) spatial dimensions of the upper layer output by a factor of 2.
The mixing block $f^{\text{mix}}_{p}: \mathbb{R}^{128 \times H_{p} \times W_{p} \times H_{\epsilon} \times W_{\epsilon}} \xrightarrow{} \mathbb{R}^{128 \times H_{p} \times W_{p} \times H_{\epsilon} \times W_{\epsilon}}$ then processes this mixture with 4D convolutions to propagate relevant information to lower layers in a top-down fashion.
After the iterative propagation, the output tensor of the lowest mixing block $f_{1}^{\text{mix}}$ is further compressed by average-pooling its last two (support) spatial dimensions, which in turn provides a 2-dimensional feature map $\mathbf{Z} \in \mathbb{R}^{128 \times H_{1} \times W_{1}}$ that signifies a condensed representation of the hypercorrelation $\mathcal{C}$.

\subsection{2D-convolutional context decoder}
\label{sec:decoder}
The decoder network consists of a series of 2D convolutions, ReLU, and upsampling layers followed by softmax function as illustrated in Fig.~\ref{fig:architecture}.
The network takes the context representation $\mathbf{Z}$ and predicts two-channel map $\mathbf{\hat{M}}^{\mathrm{q}} \in [0,1]^{2 \times H \times W}$ where two channel values indicate probabilities of foreground and background.
During training, the network parameters are optimized using the mean of cross-entropy loss between the prediction $\mathbf{\hat{M}}^{\mathrm{q}}$ and the ground-truth $\mathbf{M}^{\mathrm{q}}$ over all pixel locations.
During testing, we take the maximum channel value at each pixel to obtain final query mask prediction $\mathbf{\bar{M}}^{\mathrm{q}} \in \{0, 1\}^{H \times W}$ for evaluation.

\begin{figure*}
    \begin{center}
        \includegraphics[width=0.95\linewidth]{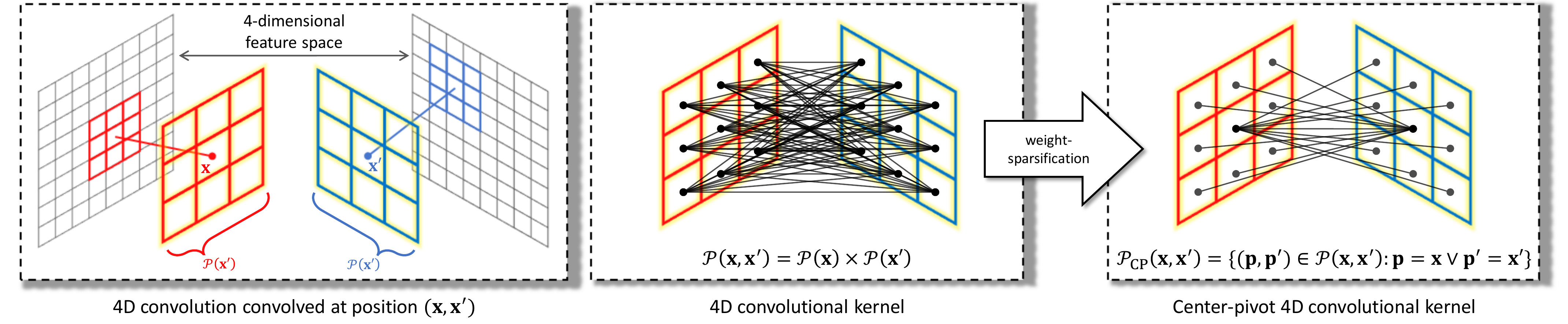}
    \end{center}
    \vspace{-5.0mm} 
      \caption{4D convolution (left) and weights of 4D kernel~\cite{rocco2018neighbourhood,yang2019optical} (middle) and center-pivot 4D kernel (right). Each black wire that connects two different pixel locations represents a single weight of the 4D kernel. The kernel size used in this example is $(3, 3, 3, 3)$, \ie, $\hat{k} = 3$.}
    \vspace{-5.0mm} 
\label{fig:kernel_vis}
\end{figure*}

\subsection{Center-pivot 4D convolution}
\label{sec:s4d}
Apparently, our network with such a large number of 4D convolutions demands a substantial amount of resources due to the curse of dimensionality, which constrained many visual correspondence methods~\cite{huang2019dynamic,li2020correspondence,li20dualrc,rocco2018neighbourhood,truong2020glunet} to use only a few 4D conv layers.
To address the concern, we revisit the 4D convolution operation and delve into its limitations.
Then we demonstrate how a unique weight-sparsification scheme effectively resolves the issues.

\smallbreak
\noindent \textbf{4D convolution and its limitation.} 
Typical 4D convolution parameterized by a kernel ${k} \in \mathbb{R}^{\hat{k} \times \hat{k} \times \hat{k} \times \hat{k}}$ on a correlation tensor $c \in \mathbb{R}^{H \times W \times H \times W}$ at position $(\mathbf{x}, \mathbf{x}') \in \mathbb{R}^{4}$\footnote{The correlation tensor $c$ is the output of cosine similarity (Eqn.~\ref{eqn:cosine}) between a pair of feature maps, $\mathbf{F}, \mathbf{F}' \in \mathbb{R}^{H \times W}$, and $\mathbf{x}$ and $\mathbf{x}'$ denote 2-dimensional spatial positions of the respective feature maps.} is formulated as
\begin{align}
\label{eqn:4d}
    ({c} * {k})(\mathbf{x}, \mathbf{x}') &= \sum_{(\mathbf{p}, \mathbf{p}') \in \mathcal{P}(\mathbf{x}, \mathbf{x}')} {c}(\mathbf{p}, \mathbf{p}') {k}(\mathbf{p} - \mathbf{x}, \mathbf{p}' - \mathbf{x}'),
\end{align}
where $\mathcal{P}(\mathbf{x}, \mathbf{x}')$ denotes a set of neighbourhood regions within the local 4D window centered on position $(\mathbf{x}, \mathbf{x}')$, \ie, $\mathcal{P}(\mathbf{x}, \mathbf{x}') = \mathcal{P}(\mathbf{x}) \times \mathcal{P}(\mathbf{x}')$ as visualized in Fig.~\ref{fig:kernel_vis}.
Although the use of 4D convolutions on a correlation tensor has shown its efficacy with good empirical performance in correspondence-related domains~\cite{huang2019dynamic,li2020correspondence,li20dualrc,rocco2018neighbourhood,truong2020glunet}, its quadratic complexity with respect to the size of input features still remains a primary bottleneck.
Another limiting factor is over-parameterization of the high-dimensional kernel:
Consider a single activation in an $n$D tensor convolved by $n$D conv kernel.
The number of times that the kernel processes this activation is exponentially proportional to $n$.
This implies some unreliable input activations with large magnitudes may entail some noise in capturing reliable patterns as a result of their excessive exposure to the high-dimensional kernel.
The work of~\cite{yang2019optical} resolves the former problem (quadratic complexity) using spatially separable 4D kernels to approximate the 4D conv with two separate 2D kernels along with additional batch normalization layers~\cite{ioffe2015batchnorm} that settle the latter problem (numerical instability).
In this work we introduce a novel weight-sparsification scheme to address both issues at the same time.

\smallbreak
\noindent \textbf{Center-pivot 4D convolution.} 
Our goal is to design a light-weight 4D kernel that is efficient in terms of both memory and time while effectively approximating the existing ones~\cite{rocco2018neighbourhood,yang2019optical}.
We achieve this via a reasonable weight-sparsification; 
from a set of neighborhood positions within a local 4D window of interest, our kernel aims to disregard a large number of activations located at fairly insignificant positions in the 4D window, thereby focusing on a small subset of relevant activations only.
Specifically, we consider the activations at positions that {\em \textbf{pivots}} either one of 2-dimensional {\em \textbf{centers}}, \eg, $\mathbf{x}$ or $\mathbf{x}'$, as the foremost influential ones as illustrated in Fig.~\ref{fig:kernel_vis}.
Given 4D position $(\mathbf{x}, \mathbf{x}')$, we collect its neighbors if and only if they are adjacent to either $\mathbf{x}$ or $\mathbf{x}'$ in its corresponding 2D subspace and define two respective sets as $\mathcal{P}_{c}(\mathbf{x}, \mathbf{x}') = \{(\mathbf{p}, \mathbf{p}') \in \mathcal{P}(\mathbf{x}, \mathbf{x}'): \mathbf{p} = \mathbf{x}\}$ and $\mathcal{P}_{c'}(\mathbf{x}, \mathbf{x}') = \{(\mathbf{p}, \mathbf{p}') \in \mathcal{P}(\mathbf{x}, \mathbf{x}'): \mathbf{p}' = \mathbf{x}'\}$.
The set of {\em center-pivot neighbours} is defined as $\mathcal{P}_{\text{CP}}(\mathbf{x}, \mathbf{x}') = \mathcal{P}_{c}(\mathbf{x}, \mathbf{x}') \cup \mathcal{P}_{c'}(\mathbf{x}, \mathbf{x}')$.
Based on these two subsets of neighbors, center-pivot 4D convolution can be formulated as a union of two separate 4D convolutions:
\begin{align}
\label{eqn:cc4d}
    (c * k_{\text{CP}})(\mathbf{x}, \mathbf{x}') &= (c * k_{c})(\mathbf{x}, \mathbf{x}') + (c * k_{c'})(\mathbf{x}, \mathbf{x}')
\end{align}
where $k_{c}$ and $k_{c'}$ are 4D kernels convolved on $\mathcal{P}_{c}(\mathbf{x}, \mathbf{x}')$ and $\mathcal{P}_{c'}(\mathbf{x}, \mathbf{x}')$ respectively.
Note that $(c * k_{c})(\mathbf{x}, \mathbf{x}')$ is equivalent to convolutions with a 2D kernel $k_{c}^{\mathrm{2D}}=k(\mathbf{0}, :) \in \mathbb{R}^{\hat{k} \times \hat{k}}$ performed on 2D slice of 4D tensor $c(\mathbf{x}, \ :)$.
Similarly, with $k_{c'}^{\mathrm{2D}}=k(:, \mathbf{0}) \in \mathbb{R}^{\hat{k} \times \hat{k}}$, we reformulate Eqn.~\ref{eqn:cc4d} as follows
\begin{align}
\label{eqn:decomposition}
    (c * k_{\text{CP}})(\mathbf{x}, \mathbf{x}') = \sum_{\mathbf{p}' \in \mathcal{P}(\mathbf{x}')} &{c}(\mathbf{x}, \mathbf{p}') {k}_{c}^{\textrm{2D}}(\mathbf{p}' - \mathbf{x}') \\ \nonumber
    + \sum_{\mathbf{p} \in \mathcal{P}(\mathbf{x})} &{c}(\mathbf{p}, \mathbf{x}') {k}_{c'}^{\textrm{2D}}( \mathbf{p} - \mathbf{x}),
\end{align}
which performs two different convolutions on separate 2D subspaces, having a linear complexity.
In Sec.~\ref{sec:ablation}, we experimentally demonstrate the superiority of the center-pivot 4D kernels over the existing ones~\cite{rocco2018neighbourhood,yang2019optical} in terms of accuracy, memory, and time.
We refer the readers to the Appendix \ref{sec:appendix_derivation} for a complete derivation of Eqn.~\ref{eqn:decomposition}.

\subsection{Extension to $K$-shot setting}
\label{sec:extension_kshot}
Our network can be easily extended to $K$-shot setting:
Given $K$ support image-mask pairs $\mathcal{S} = \{({I}^{\mathrm{s}}_{k}, \mathbf{M}^{\mathrm{s}}_{k})\}^{K}_{k=1}$and a query image ${I}^{\mathrm{q}}$, model performs $K$ forward passes to provide a set of $K$ mask predictions $\{\mathbf{\bar{M}}_{k}^{\mathrm{q}}\}_{k=1}^{K}$.
We perform voting at every pixel location by summing all the $K$ predictions and divide each output score by the maximum voting score. 
We assign foreground labels to pixels if their values are larger than some threshold $\tau$ whereas the others are classified as background.
We set $\tau = 0.5$ in our experiments.

\begin{table*}
    \begin{center}
    \scalebox{0.8}{
    \begin{tabular}{cl|cccccc|cccccc|c}
            \toprule
            \multirow{2}{*}{\shortstack{Backbone\\network}} & \multirow{2}{*}{Methods} & \multicolumn{6}{c|}{1-shot} & \multicolumn{6}{c|}{5-shot} & \# learnable \\ 
            
            & & $5^{0}$ & $5^{1}$ & $5^{2}$ & $5^{3}$ & mean & FB-IoU & $5^{0}$ & $5^{1}$ & $5^{2}$ & $5^{3}$ & mean & FB-IoU & params \\
            \midrule
            
            \multirow{6}{*}{VGG16~\cite{simonyan2015vgg}} & OSLSM~\cite{shaban2017oslsm}        & 33.6 & 55.3 & 40.9 & 33.5 & 40.8 & 61.3 & 35.9 & 58.1 & 42.7 & 39.1 & 43.9 & 61.5 & 276.7M \\  
            
            & co-FCN~\cite{rakelly2018cofcn}      & 36.7 & 50.6 & 44.9 & 32.4 & 41.1 & 60.1 & 37.5 & 50.0 & 44.1 & 33.9 & 41.4 & 60.2 & 34.2M \\  
            
            & AMP-2~\cite{siam2019amp}         & 41.9 & 50.2 & 46.7 & 34.7 & 43.4 & 61.9 & 40.3 & 55.3 & 49.9 & 40.1 & 46.4 & 62.1 & 15.8M \\
            
            & PANet~\cite{wang2019panet}       & 42.3 & 58.0 & 51.1 & 41.2 & 48.1 & 66.5 & 51.8 & 64.6 & \underline{59.8} & 46.5 & 55.7 & 70.7 & 14.7M \\ 
            
            & PFENet~\cite{tian2020pfenet}  & \underline{56.9} & \textbf{68.2} & \underline{54.4} & \underline{52.4} & \underline{58.0} & \underline{72.0} & \underline{59.0} & \textbf{69.1} & 54.8 & \underline{52.9} & \underline{59.0} & \underline{72.3} & \underline{10.4M} \\ \cline{2-15} \\[-2.0ex]
            
            & HSNet (ours)    & \textbf{59.6} & \underline{65.7} & \textbf{59.6} & \textbf{54.0} & \textbf{59.7} & \textbf{73.4} & \textbf{64.9} & \underline{69.0} & \textbf{64.1} & \textbf{58.6} & \textbf{64.1} & \textbf{76.6} & \textbf{2.6M} \\
            
            \midrule
            
            \multirow{6}{*}{ResNet50~\cite{he2016deep}} & PANet~\cite{wang2019panet}       & 44.0 & 57.5 & 50.8 & 44.0 & 49.1 & - & 55.3 & 67.2 & 61.3 & 53.2 & 59.3 & - & 23.5M \\  
            
            & PGNet~\cite{zhang2019pgnet}       & 56.0 & 66.9 & 50.6 & 50.4 & 56.0 & 69.9 & 57.7 & 68.7 & 52.9 & 54.6 & 58.5 & 70.5 & 17.2M \\  
            
            & PPNet~\cite{liu2020ppnet}      & 48.6 & 60.6 & 55.7 & 46.5 & 52.8 & 69.2 & 58.9 & 68.3 & 66.8 & 58.0 & 63.0 & \underline{75.8} & 31.5M \\  
            
            & PFENet~\cite{tian2020pfenet}  & \underline{61.7} & \underline{69.5} & 55.4 & \underline{56.3} & \underline{60.8} & \underline{73.3} & 63.1 & 70.7 & 55.8 & 57.9 & 61.9 & 73.9 & \underline{10.8M} \\ 
            
            & RePRI~\cite{malik2020repri}  & 59.8 & 68.3 & \textbf{62.1} & 48.5 & 59.7 & - & \underline{64.6} & \underline{71.4} & \textbf{71.1} & \underline{59.3} & \underline{66.6} & - & - \\ \cline{2-15} \\[-2.0ex]
            
            & HSNet (ours)    & \textbf{64.3} & \textbf{70.7} & \underline{60.3} & \textbf{60.5} & \textbf{64.0} & \textbf{76.7} & \textbf{70.3} & \textbf{73.2} & \underline{67.4} & \textbf{67.1} & \textbf{69.5} & \textbf{80.6} & \textbf{2.6M} \\

            \midrule
            
            \multirow{7}{*}{ResNet101~\cite{he2016deep}} & FWB~\cite{nguyen2019fwb}        & 51.3 & 64.5 & 56.7 & 52.2 & 56.2 & - & 54.8 & 67.4 & 62.2 & 55.3 & 59.9 & - & 43.0M \\
            
            & PPNet~\cite{liu2020ppnet}     & 52.7 & 62.8 & 57.4 & 47.7 & 55.2 & 70.9 & 60.3 & 70.0 & \textbf{69.4} & \underline{60.7} & 65.1 & \underline{77.5} & 50.5M \\  
            
            & DAN~\cite{wang2020dan}        & 54.7 & 68.6 & 57.8 & 51.6 & 58.2 & 71.9 & 57.9 & 69.0 & 60.1 & 54.9 & 60.5 & 72.3 & - \\ 
            
            & PFENet~\cite{tian2020pfenet}  & 60.5 & 69.4 & 54.4 & 55.9 & 60.1 & \underline{72.9} & 62.8 & 70.4 & 54.9 & 57.6 & 61.4 & 73.5 & \underline{10.8M} \\
            
            & RePRI~\cite{malik2020repri}  & 59.6 & 68.6 & \textbf{62.2} & 47.2 & 59.4 & - & 66.2 & 71.4 & \underline{67.0} & 57.7 & \underline{65.6} & - & - \\ \cline{2-15} \\[-2.0ex]
            
            & HSNet (ours)    & \textbf{67.3} & \textbf{72.3} & \underline{62.0} & \textbf{63.1} & \textbf{66.2} & \textbf{77.6} & \textbf{71.8} & \textbf{74.4} & \underline{67.0} & \textbf{68.3} & \textbf{70.4} & \textbf{80.6} & \textbf{2.6M} \\
            
            & HSNet$^{\dagger}$ (ours)   & \underline{66.2} & \underline{69.5} & 53.9 & \underline{56.2} & \underline{61.5} & 72.5 & \underline{68.9} & \underline{71.9} & 56.3 & {57.9} & 63.7 & 73.8 & \textbf{2.6M} \\
            
            \bottomrule
    \end{tabular}
    }
    \vspace{-2.0mm}
    \caption{\label{tab:pascal_sota}Performance on PASCAL-5$^{i}$~\cite{shaban2017oslsm} in mIoU and FB-IoU. Some results are from~\cite{malik2020repri,liu2020ppnet,tian2020pfenet,wang2020dan,yang2020pmm}. Superscript $\dagger$ denotes our model {\em \textbf{without}} support feature masking (Eqn.~\ref{eqn:masking}). Numbers in bold indicate the best performance and underlined ones are the second best.}
    \vspace{-5.0mm}
    \end{center}
\end{table*}

\begin{table*}
    \parbox{.66\linewidth}{
    \centering
        \scalebox{0.6}{
        \begin{tabular}{cl|cccccc|cccccc}
                \toprule
                \multirow{2}{*}{\shortstack{Backbone\\network}} & \multirow{2}{*}{Methods} & \multicolumn{6}{c|}{1-shot} & \multicolumn{6}{c}{5-shot} \\ 
                
                & & $20^{0}$ & $20^{1}$ & $20^{2}$ & $20^{3}$ & mean & FB-IoU & $20^{0}$ & $20^{1}$ & $20^{2}$ & $20^{3}$ & mean & FB-IoU \\

                \midrule
                
                \multirow{6}{*}{ResNet50~\cite{he2016deep}}  & PPNet~\cite{liu2020ppnet}        & 28.1 & 30.8 & 29.5 & 27.7 & 29.0 & - & 39.0 & 40.8 & 37.1 & 37.3 & 38.5 & -  \\
                
                & PMM~\cite{yang2020pmm}        & 29.3 & 34.8 & 27.1 & 27.3 & 29.6 & - & 33.0 & 40.6 & 30.3 & 33.3 & 34.3 & -  \\ 
                
                & RPMM~\cite{yang2020pmm}        & 29.5 & 36.8 & 28.9 & 27.0 & 30.6 & - & 33.8 & 42.0 & 33.0 & 33.3 & 35.5 & -   \\    
                
                & PFENet~\cite{tian2020pfenet}        & \textbf{36.5} & 38.6 & \underline{34.5} & \underline{33.8} & \underline{35.8} & - & 36.5 & 43.3 & 37.8 & 38.4 & 39.0 & - \\   
                
                & RePRI~\cite{malik2020repri}        & 32.0 & \underline{38.7} &  32.7 & {33.1} & 34.1 & - & \underline{39.3} & \underline{45.4} & \underline{39.7} & \underline{41.8} & \underline{41.6} & - \\    \cline{2-14} \\[-2.0ex]
                
                & HSNet (ours)    & \underline{36.3} & \textbf{43.1} & \textbf{38.7} & \textbf{38.7} & \textbf{39.2} & \textbf{68.2} & \textbf{43.3} & \textbf{51.3} & \textbf{48.2} & \textbf{45.0} & \textbf{46.9} & \textbf{70.7} \\
                
                \midrule
                
                \multirow{4}{*}{ResNet101~\cite{he2016deep}} & FWB~\cite{nguyen2019fwb}        & 17.0 & 18.0 & 21.0 & 28.9 & 21.2 & - & 19.1 & 21.5 & 23.9 & 30.1 & 23.7 & -  \\ 
                
                & DAN~\cite{wang2020dan}  & - & - & - & - & 24.4 & 62.3 & - & - & - & - & 29.6 & 63.9  \\
                
                & PFENet~\cite{tian2020pfenet}  & \underline{36.8} & \underline{41.8} & \underline{38.7} & \underline{36.7} & \underline{38.5} & \underline{63.0} & \underline{40.4} & \underline{46.8} & \underline{43.2} & \underline{40.5} & \underline{42.7} & \underline{65.8} \\  \cline{2-14} \\[-2.0ex]
                
                & HSNet (ours)    & \textbf{37.2} & \textbf{44.1} & \textbf{42.4} & \textbf{41.3} & \textbf{41.2} & \textbf{69.1} & \textbf{45.9} & \textbf{53.0} & \textbf{51.8} & \textbf{47.1} & \textbf{49.5} & \textbf{72.4}  \\
                
                \bottomrule
        \end{tabular}
        }
        \vspace{-2.0mm}
        \caption{\label{tab:coco_sota}Performance on COCO-20$^{i}$~\cite{nguyen2019fwb} in mIoU and FB-IoU. The results of other methods are from~\cite{malik2020repri,liu2020ppnet,tian2020pfenet,wang2020dan,yang2020pmm}.}
        \vspace{-5.0mm}
    }
    \hfill
    \parbox{.30\linewidth}{
    \centering
        
        \scalebox{0.65}{
        \begin{tabular}{cl|cc}
                \toprule
                \multirow{2}{*}{\shortstack{Backbone\\network}} & \multirow{2}{*}{Methods} & \multicolumn{2}{c}{mIoU} \\ 
                
                & & 1-shot & 5-shot \\
                
                \midrule
                
                \multirow{5}{*}{VGG16~\cite{simonyan2015vgg}} & OSLSM~\cite{shaban2017oslsm} & 70.3 & 73.0 \\
                
                & GNet~\cite{rakelly2018gnet} & 71.9 & 74.3 \\
                 
                & FSS~\cite{li2020fss1000} & 73.5 & 80.1 \\
                 
                & DoG-LSTM~\cite{azad2021doglstm} & \underline{80.8} & \underline{83.4} \\  \cline{2-4} \\[-2.0ex]
                 
                & HSNet (ours) & \textbf{82.3} & \textbf{85.8} \\
                \midrule
                ResNet50~\cite{he2016deep}      & HSNet (ours)  & \textbf{85.5} & \textbf{87.8} \\
                \midrule
                \multirow{2}{*}{ResNet101~\cite{he2016deep}} & DAN~\cite{wang2020dan} & \underline{85.2} & \underline{88.1} \\  \cline{2-4} \\[-2.0ex]
                
                & HSNet (ours)  & \textbf{86.5} & \textbf{88.5} \\
                
                \bottomrule
        \end{tabular}
        }
        \vspace{-2.0mm}
        \caption{\label{tab:fss_sota}Mean IoU comparison on FSS-1000~\cite{li2020fss1000}. Some results are from~\cite{azad2021doglstm,wang2020dan}.}
        \vspace{-5.0mm}
    }
\end{table*}

\section{Experiment}
In this section we evaluate the proposed method, compare it with recent state of the arts, and provide in-depth analyses of the results with ablation study.

\smallbreak
\noindent \textbf{Implementation details.}
For the backbone network, we employ VGG~\cite{simonyan2015vgg} and ResNet~\cite{he2016deep} families pre-trained on ImageNet~\cite{deng2009imagenet}, \eg, VGG16, ResNet50, and ResNet101.
For VGG16 backbone, we extract features after every conv layer in the last two building blocks: from \texttt{conv4\_x} to \texttt{conv5\_x}, and after the last maxpooling layer.
For ResNet backbones, we extract features at the end of each bottleneck before ReLU activation: from \texttt{conv3\_x} to \texttt{conv5\_x}.
This feature extracting scheme results in 3 pyramidal layers ($P=3$) for each backbone.
We set spatial sizes of both support and query images to $400 \times 400$, \ie, $H,W = 400$, thus having $H_{1},W_{1} = 50$, $H_{2},W_{2} = 25$, and $H_{3},W_{3} = 13$.
The network is implemented in PyTorch~\cite{pytorch} and optimized using Adam~\cite{kingma2015adam} with learning rate of 1e-3.
We freeze the pre-trained backbone networks to prevent them from learning class-specific representations of the training data.

\smallbreak
\noindent \textbf{Datasets.} We evaluate the proposed network on three standard few-shot segmentation datasets: PASCAL-5${^i}$~\cite{shaban2017oslsm}, COCO-20${^i}$~\cite{lin2015coco}, and FSS-1000~\cite{li2020fss1000}.
PASCAL-5${^i}$ is created from PASCAL VOC 2012~\cite{everingham2015pascal} with extra mask annotations~\cite{hariharan2014sds}, consisting of 20 object classes that are evenly divided into 4 folds: $\{5^{i}: i \in \{0,1,2,3\}\}$.
COCO-20${^i}$ consists of mask-annotated images from 80 object classes divided into 4 folds: $\{20^{i}: i \in \{0,1,2,3\}\}$.
Following common training/evaluation scheme~\cite{liu2020ppnet,nguyen2019fwb,tian2020pfenet,wang2020dan,yang2020pmm}, we conduct cross-validation over all the folds;
for each fold $i$, samples from the other remaining folds are used for training and 1,000 episodes from the target fold $i$ are randomly sampled for evaluation. 
For every fold, we use the same model with the same hyperparameter setup following the standard cross-validation protocol.  
FSS-1000 contains mask-annotated images from 1,000 classes divided into training, validation and test splits having 520, 240, and 240 classes respectively.

\smallbreak
\noindent \textbf{Evaluation metrics.}
We adopt mean intersection over union (mIoU) and foreground-background IoU (FB-IoU) as our evaluation metrics. 
The mIoU metric averages over IoU values of all classes in a fold: $\text{mIoU} = \frac{1}{C}\sum_{c=1}^{C}\text{IoU}_{c}$ where $C$ is the number of classes in the target fold and $\text{IoU}_{c}$ is the intersection over union of class $c$.
FB-IoU ignores object classes and computes average of foreground and background IoUs: $\text{FB-IoU} = \frac{1}{2}(\text{IoU}_{\mathrm{F}} + \text{IoU}_{\mathrm{B}})$ where $\text{IoU}_{\mathrm{F}}$ and $\text{IoU}_{\mathrm{B}}$ are respectively foreground and background IoU values in the target fold.
As mIoU better reflects model generalization capability and prediction quality than FB-IoU does, we mainly focus on mIoU in our experiments.

\subsection{Results and analysis}
\label{sec:results_analyses}
We evaluate the proposed model on PASCAL-5$^{i}$, COCO-20$^{i}$, and FSS-1000 and compare the results with recent methods~\cite{malik2020repri,liu2020ppnet,nguyen2019fwb,rakelly2018cofcn,shaban2017oslsm,siam2019amp,tian2020pfenet,wang2020dan,wang2019panet,zhang2019pgnet}.
Table~\ref{tab:pascal_sota} summarizes 1-shot and 5-shot results on PASCAL-5$^{i}$;
all of our models with three different backbones clearly set new state of the arts with the smallest the number of learnable parameters.
With ResNet101 backbone, our 1-shot and 5-shot results respectively achieve 6.1\%p and 4.8\%p of mIoU improvements over \cite{tian2020pfenet} and \cite{malik2020repri}, verifying its superiority in few-shot segmentation task.
As shown in Tab.~\ref{tab:coco_sota}, our model outperforms recent methods with a sizable margin on COCO-20$^{i}$ as well, achieving 2.7\%p (1-shot) and 6.8\%p (5-shot) of mIoU improvements over \cite{tian2020pfenet} with ResNet101 backbone.
Also on the last benchmark, FSS-1000, our method sets a new state of the art, outperforming~\cite{azad2021doglstm,wang2020dan} as shown in Tab.~\ref{tab:fss_sota}. 

We conduct additional experiments without support feature masking (Eqn.~\ref{eqn:masking}).
Note that this setup is similar to co-segmentation problem~\cite{chen2020show, taniai2016joint, yang2014daisy} with stronger demands for generalizibility since the model is evaluated on novel classes.
As seen in the bottom row of Tab.~\ref{tab:pascal_sota}, our model without support masking still performs remarkably well, achieving 1.4\%p mIoU improvement over the previous best method~\cite{tian2020pfenet} in 1-shot setting whereas it rivals \cite{malik2020repri,tian2020pfenet} in 5-shot setting.
This interesting result reveals that our model is also capable of identifying `common' instances across different input images as well as predicting fine-grained segmentation masks.

\smallbreak
\noindent \textbf{Robustness to domain shift.}
To demonstrate the robustness of our method to domain shift, we evaluate COCO-trained HSNet on each fold of PASCAL-5$^{i}$ following the recent work of~\cite{malik2020repri}.
We use the same training/test folds as in~\cite{malik2020repri} where object classes in training and testing do not overlap.
As seen in Tab.~\ref{tab:domain_shift}, our model, which is {\em trained without any data augmentation methods} with 18 times smaller number of trainable parameters compared to~\cite{malik2020repri} (2.6M {\em vs.}~46.7M),  performs robustly in presence of large domain gaps between COCO-20$^{i}$ and PASCAL-5$^{i}$, surpassing~\cite{malik2020repri} by 1.0\%p in 5-shot setting, and further improves with a larger backbone, \eg, ResNet101. 
The results clearly show the robustness of our method to domain shift, and may further increase when trained with data augmentations used in~\cite{malik2020repri, tian2020pfenet}.

\begin{table}[t]
    \begin{center}
    \scalebox{0.7}{
    \begin{tabular}{lcccc}
            \toprule
            \multirow{2}{*}{Method} & \multicolumn{2}{c}{COCO$\rightarrow$PASCAL} & \# params & data augmentation \\
            & 1-shot & 5-shot & to train & used during training \\
            
            \midrule
            
            PFENet$_{\text{res50}}$~\cite{tian2020pfenet} & 61.1 & 63.4 & \underline{10.8M} & flip, rotate, crop \\
            RePRI$_{\text{res50}}$~\cite{malik2020repri} & \textbf{63.2} & \underline{67.7} & 46.7M & flip \\
            
            \midrule
            
            HSNet$_{\text{res50}}$(ours) & \underline{61.6} & \textbf{68.7} & \textbf{2.6M} & \textbf{none} \\
            \midrule
            \midrule
            HSNet$_{\text{res101}}$(ours) & \textbf{64.1} & \textbf{70.3} & \textbf{2.6M} & \textbf{none} \\
            
            \bottomrule
    \end{tabular}
    }
    \vspace{-2.0mm}
    \caption{\label{tab:domain_shift}\small{Domain shift results. Subscripts denote backbone.}}
    \vspace{-10.0mm}
    \end{center}
\end{table}

\subsection{Ablation study}
\label{sec:ablation}
We conduct extensive ablation study to investigate the impacts of major components in our model: hypercorrelations, pyramidal architecture, and center-pivot 4D kernels.
We also study how freezing backbone networks prevents overfitting and helps generalization on novel classes.
All ablation study experiments are performed with ResNet101 backbone on PASCAL-5$^{i}$~\cite{shaban2017oslsm} dataset.

\begin{figure}[t]
    \begin{center}
        \includegraphics[width=0.99\linewidth]{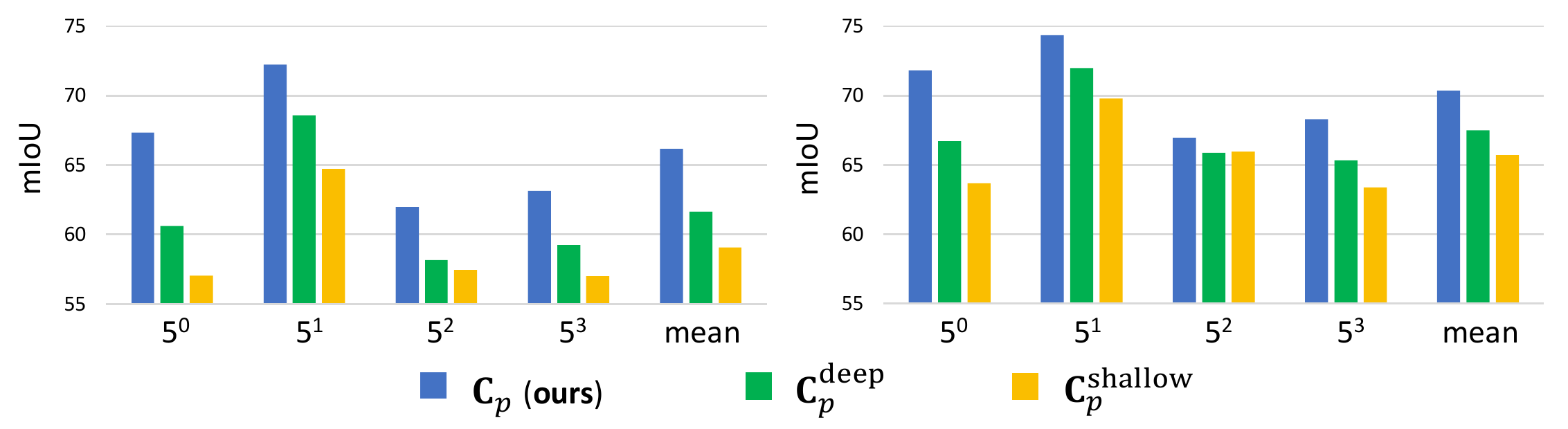}
    \end{center}
    \vspace{-3.0mm} 
        \caption{Ablation study on hypercorrelations on PASCAL-5$^{i}$~\cite{shaban2017oslsm} dataset in 1-shot (left) and 5-shot (right) mIoU results.}
    \vspace{-3.0mm} 
    \label{fig:hypercorr_ablation_quant}
\end{figure}

\smallbreak
\noindent \textbf{Ablation study on hypercorrelations.} 
To study the effect of intermediate correlations $\{\mathbf{\hat{C}}_{l}\}_{l \in \mathcal{L}_{p}}$ in hypercorrelation $\mathbf{C}_{p} \in \mathbb{R}^{|\mathcal{L}_{p}| \times H_{p} \times W_{p} \times H_{p} \times W_{p}}$, we form single-channel hypercorrelations using only {\em a single intermediate correlation}.
Specifically, we form two different single-channel hypercorrelations using the smallest (shallow) and largest (deep) layer indices in $\mathcal{L}_{p}$ and denote the hypercorrelations as $\mathbf{C}_{p}^{\text{shallow}}, \mathbf{C}_{p}^{\text{deep}} \in \mathbb{R}^{1 \times H_{p} \times W_{p} \times H_{p} \times W_{p}}$, and compare the results with ours ($\mathbf{C}_{p}$) in Fig.~\ref{fig:hypercorr_ablation_quant}.
The large performance gaps between $\mathbf{C}_{p}$ and the single-channel hypercorrelations confirm that capturing diverse correlation patterns from dense intermediate CNN layers is crucial in effective pattern analyses.
Performance degradation from $\mathbf{C}_{p}^{\text{deep}}$ to $\mathbf{C}_{p}^{\text{shallow}}$ indicates that reliable feature representations typically appear at deeper layers of a CNN.

\begin{figure}[t]
    \begin{center}
        \includegraphics[width=0.99\linewidth]{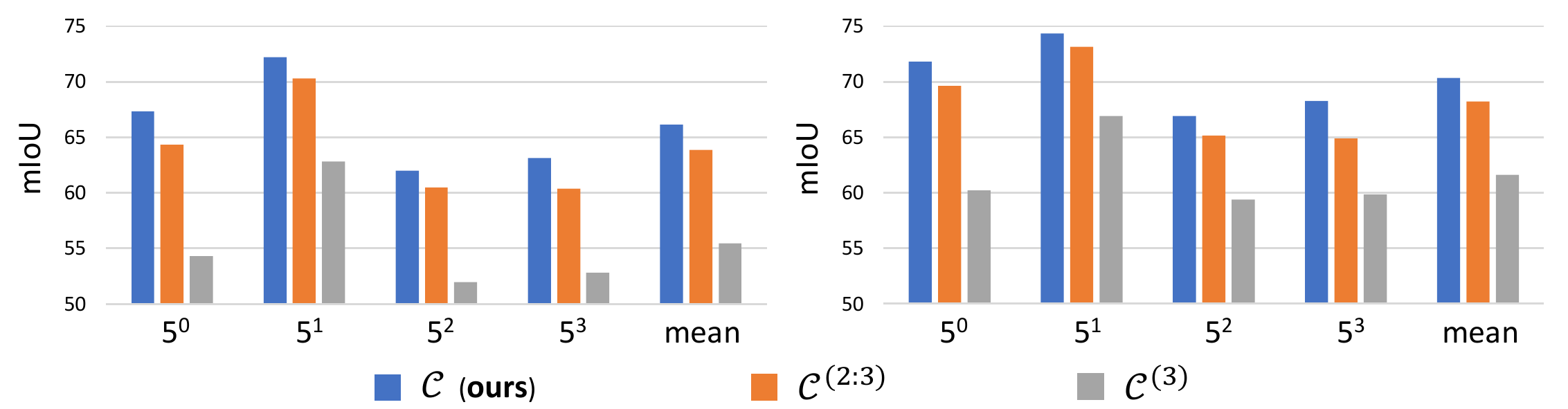}
    \end{center}
    \vspace{-3.0mm} 
        \caption{Ablation study on pyramid layers on PASCAL-5$^{i}$~\cite{shaban2017oslsm} dataset in 1-shot (left) and 5-shot (right) mIoU results.}
    \vspace{-3.0mm} 
    \label{fig:pyramid_ablation_quant}
\end{figure}

\begin{figure}[t]
    \begin{center}
        \includegraphics[width=0.99\linewidth]{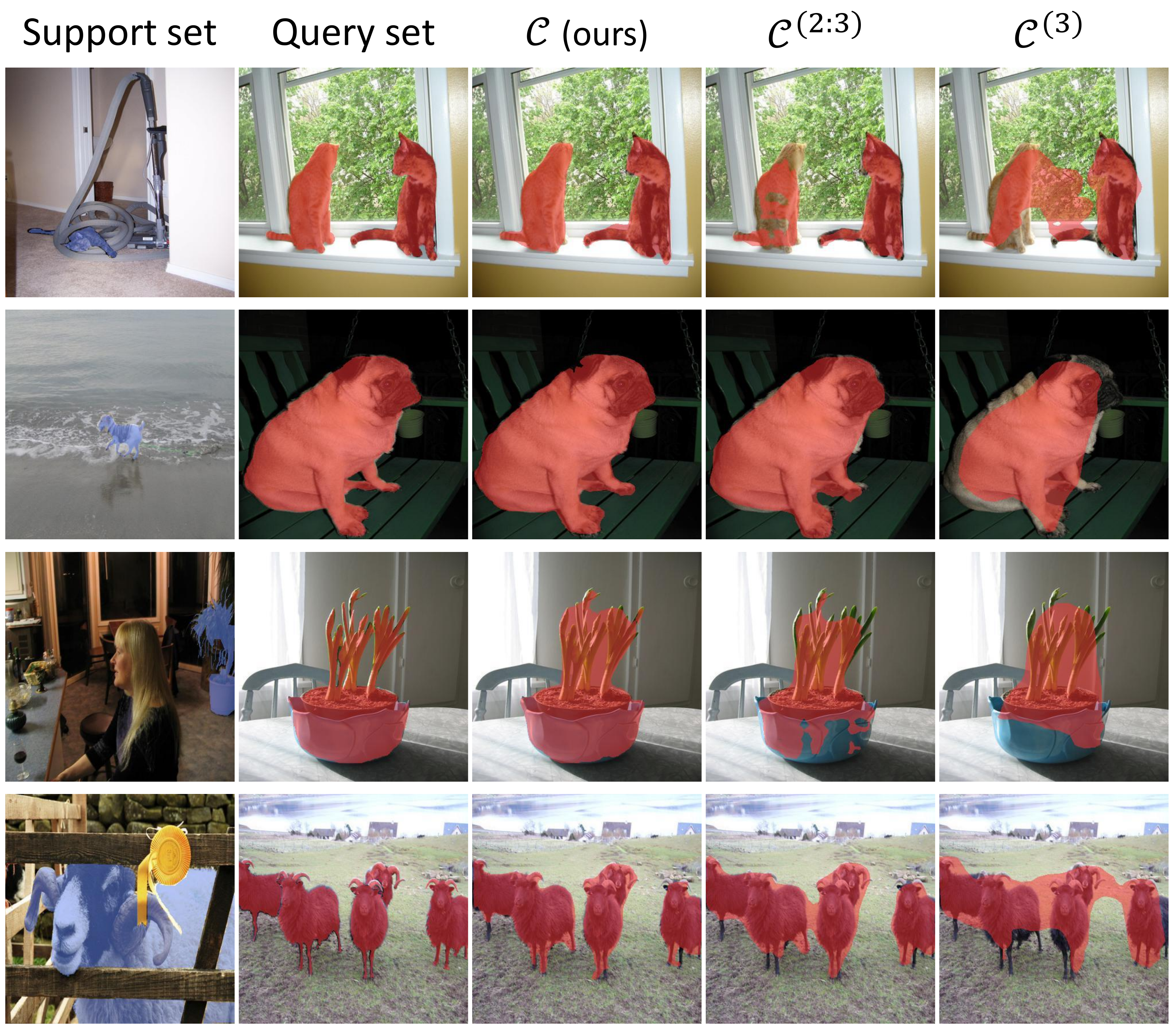}
    \end{center}
    \vspace{-5.0mm} 
    \caption{Ablation study on hypercorrelation pyramid layers.}
    \vspace{-6.0mm} 
    \label{fig:pyramid_ablation_qual}
\end{figure}

\begin{table*}
    \begin{center}
    \scalebox{0.72}{
    \begin{tabular}{l|ccccc|ccccc|cccc}
            \toprule
            
            \multirow{2}{*}{Kernel type} & \multicolumn{5}{c|}{1-shot} & \multicolumn{5}{c|}{5-shot} & \multirow{2}{*}{\shortstack{\# learnable\\params}} & time & memory footprint & FLOPs \\
            
            & 5$^{0}$ & 5$^{1}$ & 5$^{2}$ & 5$^{3}$ & mean & 5$^{0}$ & 5$^{1}$ & 5$^{2}$ & 5$^{3}$ & mean & & ({\em ms}) & ({\em GB}) & ({\em G}) \\
            
            \midrule
            
            Original 4D kernel~\cite{rocco2018neighbourhood} & 64.5 & 71.4 & \underline{62.3} & 61.7 & 64.9 & 70.8 & \textbf{74.8} & \textbf{67.4} & 67.5 & 70.1 & 11.3M & 512.17 & 4.12 & 702.35 \\
            
            Separable 4D kernel~\cite{yang2019optical} & \underline{66.1} & \underline{72.0} & \textbf{63.2} & \underline{62.6} & \underline{65.9} & \underline{71.2} & 74.1 & \underline{67.2} & \underline{68.1} & \underline{70.2} & \underline{4.4M} & \underline{28.48} & \underline{1.50} & \underline{28.40} \\
            
            \midrule
            
            Center-pivot 4D kernel (ours) & \textbf{67.3} & \textbf{72.3} & 62.0 & \textbf{63.1} & \textbf{66.2} & \textbf{71.8} & \underline{74.4} & 67.0 & \textbf{68.3} & \textbf{70.4} & \textbf{2.6M} & \textbf{25.51} & \textbf{1.39} & \textbf{20.56} \\
            
            \bottomrule
    \end{tabular}
    }
    \vspace{-2.0mm}
    \caption{\label{tab:kernel_comp}Comparison between three different 4D conv kernels in model size, per-episode inference time, memory consumption and FLOPs. For fair comparison, the inference times of all the models are measured on a machine with an Intel i7-7820X and an NVIDIA Titan-XP.}
    \vspace{-9.0mm}
    \end{center}
\end{table*}

\smallbreak
\noindent \textbf{Ablation study on pyramid layers.} 
To see the impact of hypercorrelation $\mathbf{C}_{p}$ at each layer $p$, we perform experiments in absence of each pyramidal layer.
We train and evaluate our model using two different hypercorrelation pyramids, $\mathcal{C}^{(2:3)}=\{\mathbf{C}_{2}, \mathbf{C}_{3}\}$ and $\mathcal{C}^{(3)}=\{\mathbf{C}_{3}\}$, and compare the results with ours $\mathcal{C}=\{\mathbf{C}_{p}\}_{p=1}^{3}$.
Figure~\ref{fig:pyramid_ablation_quant} summarizes the results;
given hypercorrelation pyramid without geometric information ($\mathcal{C}^{(2:3)}$), our model fails to refine object boundaries in the final mask prediction as visualized in Fig.~\ref{fig:pyramid_ablation_qual}.
Given a single hypercorrelation that only encodes semantic relations ($\mathcal{C}^{(3)}$), the model predictions are severely damaged, providing only rough localization of the target objects.
These results indicate that capturing patterns of both semantic and geometric cues is essential for fine-grained localization.

\smallbreak
\noindent \textbf{Comparison between three different 4D kernels.} 
We conduct ablation study on 4D kernel by replacing the proposed center-pivot 4D kernel with the original~\cite{rocco2018neighbourhood} and spatially separable~\cite{yang2019optical} 4D kernels and compare their model size, per-episode inference time (1-shot), memory consumption, and floating point operations per second (FLOPs) with ours.
Table~\ref{tab:kernel_comp} summarizes the results.
The proposed kernel records the fastest inference time with the smallest memory/FLOPs requirements while being comparably effective than the other two.
The results clearly support our claim that a large part of parameters in a high-dimensional kernel can safely be discarded without harming the quality of predictions; only a few relevant parameters are sufficient and even better for the purpose.
While both the separable~\cite{yang2019optical} and our center-pivot 4D convolutions operate on two separate 2D convolutions, auxiliary transformation layers with multiple batch normalizations that make the separable 4D conv numerically stable in its sequential design result in twice larger number of parameters (4.4M vs. 2.6M) and slower inference time (28.48ms vs. 25.51ms) than ours.

\begin{figure}[t]
    \centering
    \includegraphics[width=0.99\linewidth]{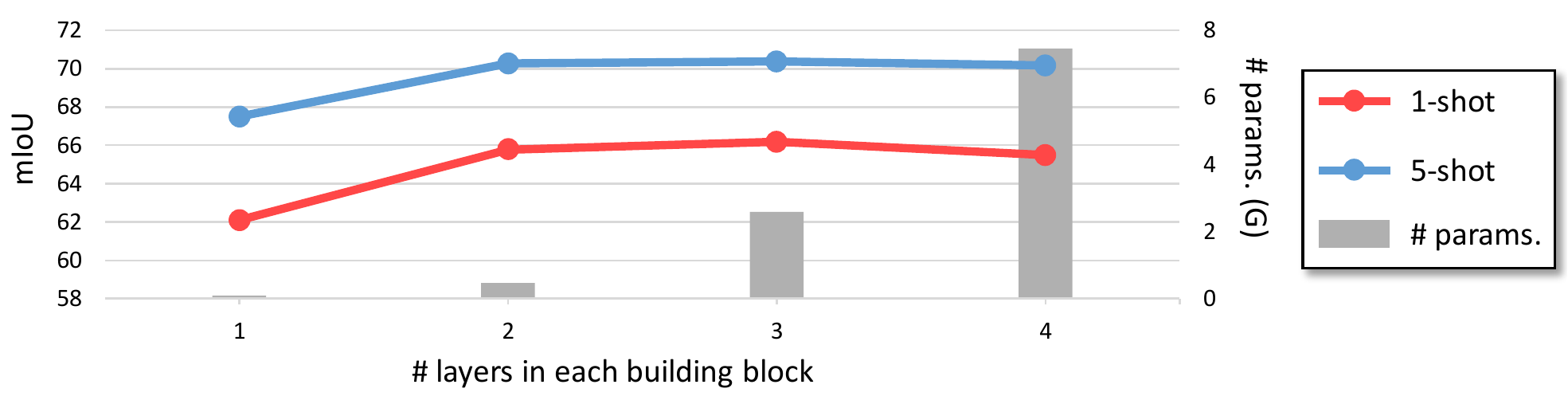}
	\vspace{-2.0mm}
	\caption{\label{fig:num_layers}The effect of depths in building blocks: $f_{p}^{\text{sqz}}$ and $f_{p}^{\text{mix}}$.}
	\vspace{-6.0mm}
\end{figure}

\smallbreak
\noindent \textbf{The number of 4D layers in building blocks.} 
We also perform experiments with varying number of 4D conv layers in the two building blocks: $f_{p}^{\text{sqz}}$ and $f_{p}^{\text{mix}}$. 
Figure~\ref{fig:num_layers} plots 1-shot and 5-shot mIoU results on PASCAL-5$^{i}$ with the model sizes.
In the experiments, appending additional 4D layers (with a group norm and a ReLU activation) in the building blocks provides clear performance improvements up to three layers but the accuracy eventually saturates after all.
Hence we use a stack of three 4D layers for both.

\begin{figure}[t]
    \centering
    \includegraphics[width=0.99\linewidth]{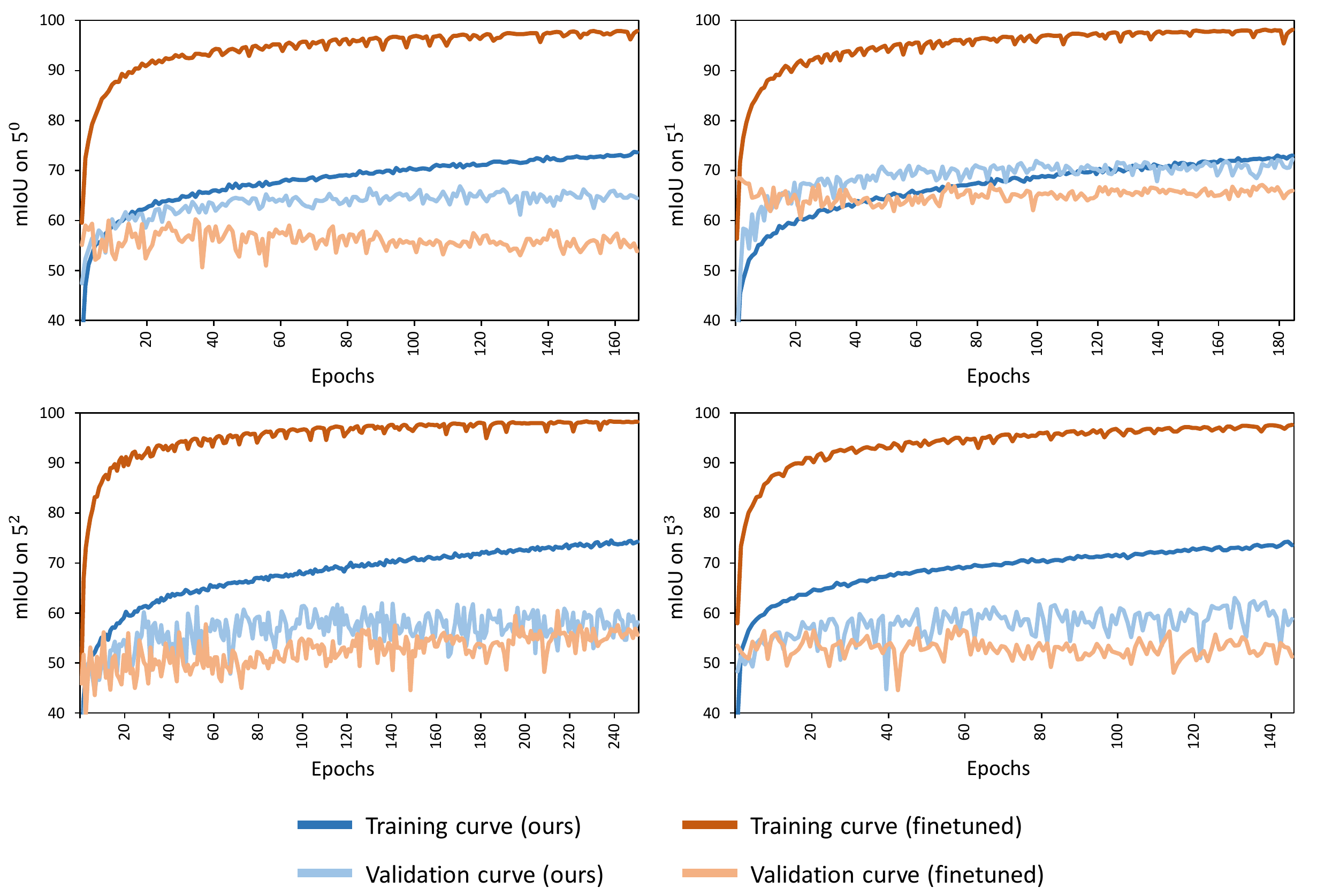}
	\vspace{-2.0mm}
	\caption{\label{fig:finetuning}Learning curves (x-axis: epoch, y-axis: mIoU) on PASCAL-5$^{i}$. We carefully tuned the learning rate of the backbone and set it to 100 times smaller than the layers in HSNet (1e-5).}
	\vspace{-5.0mm}
\end{figure}

\smallbreak
\noindent \textbf{Finetuning backbone networks.} 
To investigate the significance of learning `feature correlations' over learning `feature representation' in few-shot regime, we finetune our backbone network and compare learning processes of the finetuned model and ours (frozen backbone).
Figure~\ref{fig:finetuning} plots the training/validation curves of the finetuned model and ours on every fold of PASCAL-5$^{i}$.
The finetuned model rapidly overfits to the training data, losing generic, comprehensive visual representations learned from large-scale dataset~\cite{deng2009imagenet}.
Meanwhile, our model with frozen backbone provides better generalizibility with large trade-offs between training and validation accuracies.
The results reveal that learning new appearances under limited supervision requires understanding their `relations' to diverse visual patterns acquired from a vast amount of past experiences, \eg, ImageNet classification.
This is quite analogous to human vision perspective in the sense that we generalize novel concepts (what we see) by analyzing their relations to the past observations (what we know)~\cite{markman1989chlidren}.

For additional experimental details, results and analyses, we refer the readers to the Appendix.

\section{Conclusion}
We have presented a novel framework that analyzes complex feature correlations in a fully-convolutional manner using light-weight 4D convolutions.
The significant performance improvements on three standard benchmarks demonstrate that learning patterns of {\em feature relations} from {\em multiple visual aspects} is effective in fine-grained segmentation under limited supervision.
We also demonstrated a unique way of discarding insignificant weights leads to an efficient decomposition of a 4D kernel into a pair of 2D kernels, thus allowing extensive use of 4D conv layers at a significantly small cost.
We believe our investigation will further facilitate the use of 4D convolutions in other domains that require learning to analyze high-dimensional correlations.

\vspace{7.0mm}
\smallbreak
\noindent \textbf{Acknowledgements.}
This work was supported by Samsung Advanced Institute of Technology (SAIT), the NRF grant (NRF-2017R1E1A1A01077999), and the IITP grant (No.2019-0-01906, AI Graduate School Program - POSTECH) funded by Ministry of Science and ICT, Korea.

\clearpage

{\small
    \bibliographystyle{ieee_fullname}
    \bibliography{egbib}
}

\newcommand{\beginsupplement}{%
        \setcounter{table}{0}
        \renewcommand{\thetable}{A\arabic{table}}%
        \setcounter{figure}{0}
        \renewcommand{\thefigure}{A\arabic{figure}}%
     }
\beginsupplement


\clearpage
\begin{appendices}

\section{Complete derivation of the center-pivot 4D convolution}
\label{sec:appendix_derivation}
In this section, we extend Sec.~\ref{sec:s4d} to provide a complete derivation of the center-pivot 4D convolution.
Note that a typical 4D convolution parameterized by a kernel ${k} \in \mathbb{R}^{\hat{k} \times \hat{k} \times \hat{k} \times \hat{k}}$ on a correlation tensor $c \in \mathbb{R}^{H \times W \times H \times W}$ at position $(\mathbf{x}, \mathbf{x}') \in \mathbb{R}^{4}$ is formulated as
\begin{align}
    ({c} * {k})(\mathbf{x}, \mathbf{x}') &= \sum_{(\mathbf{p}, \mathbf{p}') \in \mathcal{P}(\mathbf{x}, \mathbf{x}')} {c}(\mathbf{p}, \mathbf{p}') {k}(\mathbf{p} - \mathbf{x}, \mathbf{p}' - \mathbf{x}'),
\end{align}
where $\mathcal{P}(\mathbf{x}, \mathbf{x}')$ denotes a set of neighbourhood regions within the local 4D window centered on position $(\mathbf{x}, \mathbf{x}')$, \ie, $\mathcal{P}(\mathbf{x}, \mathbf{x}') = \mathcal{P}(\mathbf{x}) \times \mathcal{P}(\mathbf{x}')$ as visualized in Fig.~\ref{fig:kernel_vis}.
Now we design a light-weight, efficient 4D convolution via a reasonable weight-sparsification; 
from a set of neighborhood positions within a local 4D window of interest, our kernel aims to disregard a large number of activations located at fairly insignificant positions in the 4D window, thereby focusing only on a small subset of relevant activations for capturing complex patterns in the correlation tensor.
Specifically, we consider activations at positions that {\em \textbf{pivots}} either one of 2-dimensional {\em \textbf{centers}}, \eg, $\mathbf{x}$ or $\mathbf{x}'$, as the foremost influential ones.
Given 4D position $(\mathbf{x}, \mathbf{x}')$, we collect its neighbors if and only if they are adjacent to either $\mathbf{x}$ or $\mathbf{x}'$ in its corresponding 2D subspace and define two respective sets as
\begin{align}
    \mathcal{P}_{c}(\mathbf{x}, \mathbf{x}') = \{(\mathbf{p}, \mathbf{p}') \in \mathcal{P}(\mathbf{x}, \mathbf{x}'): \mathbf{p} = \mathbf{x}\},
\end{align}
and
\begin{align}
    \mathcal{P}_{c'}(\mathbf{x}, \mathbf{x}') = \{(\mathbf{p}, \mathbf{p}') \in \mathcal{P}(\mathbf{x}, \mathbf{x}'): \mathbf{p}' = \mathbf{x}'\}.
\end{align}
The set of center-pivot neighbours $\mathcal{P}_{\text{CP}}(\mathbf{x}, \mathbf{x}')$ is defined as a union of the two subsets:
\begin{align}
    \mathcal{P}_{\text{CP}}(\mathbf{x}, \mathbf{x}') = \mathcal{P}_{c}(\mathbf{x}, \mathbf{x}') \cup \mathcal{P}_{c'}(\mathbf{x}, \mathbf{x}').
\end{align}
Based on this small subset of neighbors, center-pivot 4D (CP 4D) convolution can be formulated as a union of two separate 4D convolutions:
\begin{align}
\label{eqn:cc4d2}
    (c * k_{\text{CP}})(\mathbf{x}, \mathbf{x}') &= (c * k_{c})(\mathbf{x}, \mathbf{x}') + (c * k_{c'})(\mathbf{x}, \mathbf{x}'),
\end{align}
where $k_{c}$ and $k_{c'}$ are 4D kernels with their respective neighbours $\mathcal{P}_{c}(\mathbf{x}, \mathbf{x}')$ and $\mathcal{P}_{c'}(\mathbf{x}, \mathbf{x}')$.
Now consider below
\begin{align}
    \nonumber
    (c * k_{c})(\mathbf{x}, \mathbf{x}') &= \sum_{(\mathbf{p}, \mathbf{p}') \in \mathcal{P}_{c}(\mathbf{x}, \mathbf{x}')} {c}(\mathbf{p}, \mathbf{p}') {k}(\mathbf{p} - \mathbf{x}, \mathbf{p}' - \mathbf{x}') \\ \nonumber
    &= \sum_{\mathbf{p}' \in \mathcal{P}(\mathbf{x}')} {c}(\mathbf{x}, \mathbf{p}') {k}(\mathbf{x} - \mathbf{x}, \mathbf{p}' - \mathbf{x}') \\ \nonumber
    &= \sum_{\mathbf{p}' \in \mathcal{P}(\mathbf{x}')} {c}(\mathbf{x}, \mathbf{p}') {k}(\mathbf{0}, \mathbf{p}' - \mathbf{x}') \\ 
    &= \sum_{\mathbf{p}' \in \mathcal{P}(\mathbf{x}')} {c}(\mathbf{x}, \mathbf{p}') {k}_{c}^{\textrm{2D}}(\mathbf{p}' - \mathbf{x}'),
\end{align}
which is equivalent to convolution with kernel $k_{c}^{\mathrm{2D}}=k(\mathbf{0}, :) \in \mathbb{R}^{\hat{k} \times \hat{k}}$ performed on 2D slice of the 4D tensor $c(\mathbf{x}, \ :)$.
Similarly,
\begin{align}
    \nonumber
    (c * k_{c'})(\mathbf{x}, \mathbf{x}') &= \sum_{(\mathbf{p}, \mathbf{p}') \in \mathcal{P}_{c'}(\mathbf{x}, \mathbf{x}')} {c}(\mathbf{p}, \mathbf{p}') {k}(\mathbf{p} - \mathbf{x}, \mathbf{p}' - \mathbf{x}') \\ \nonumber
    &= \sum_{\mathbf{p} \in \mathcal{P}(\mathbf{x})} {c}(\mathbf{p}, \mathbf{x}') {k}(\mathbf{p} - \mathbf{x}, \mathbf{x}' - \mathbf{x}') \\ \nonumber
    &= \sum_{\mathbf{p} \in \mathcal{P}(\mathbf{x})} {c}(\mathbf{p}, \mathbf{x}') {k}(\mathbf{p} - \mathbf{x}, \mathbf{0}) \\ 
    &= \sum_{\mathbf{p} \in \mathcal{P}(\mathbf{x})} {c}(\mathbf{p}, \mathbf{x}') {k}_{c'}^{\textrm{2D}}(\mathbf{p} - \mathbf{x}),
\end{align}
where $k_{c'}^{\mathrm{2D}}=k(:, \mathbf{0}) \in \mathbb{R}^{\hat{k} \times \hat{k}}$.
Based on above derivations, we rewrite Eqn.~\ref{eqn:cc4d2} as follows
\begin{align}
    \nonumber
    (c * k_{\text{CP}})(\mathbf{x}, \mathbf{x}') = \sum_{\mathbf{p}' \in \mathcal{P}(\mathbf{x}')} &{c}(\mathbf{x}, \mathbf{p}') {k}_{c}^{\textrm{2D}}(\mathbf{p}' - \mathbf{x}') \\
    + \sum_{\mathbf{p} \in \mathcal{P}(\mathbf{x})} &{c}(\mathbf{p}, \mathbf{x}') {k}_{c'}^{\textrm{2D}}( \mathbf{p} - \mathbf{x}),
\end{align}
which performs two different convolutions on separate 2D subspaces, having a linear complexity.

\section{Implementation details}
\label{sec:appendix_implementation}
For the backbone networks, we employ VGG~\cite{simonyan2015vgg} and ResNet~\cite{he2016deep} families pre-trained on ImageNet~\cite{deng2009imagenet}, \eg, VGG16, ResNet50, and ResNet101.
For the VGG16 backbone, we extract features after every conv layer in the last two building blocks: from \texttt{conv4\_x} to \texttt{conv5\_x}, and after the last maxpooling layer.
For the ResNet backbones, we extract features at the end of each bottleneck before ReLU activation: from \texttt{conv3\_x} to \texttt{conv5\_x}.
This feature extracting scheme results in 3 pyramidal layers ($P=3$) for every backbone.
We set spatial sizes of both support and query images to $400 \times 400$, \ie, $H,W = 400$, thus having $H_{1},W_{1} = 50$, $H_{2},W_{2} = 25$, and $H_{3},W_{3} = 13$ for both ResNet50 and ResNet101 bakcbones and $H_{1},W_{1} = 50$, $H_{2},W_{2} = 25$, and $H_{3},W_{3} = 12$ for the VGG16 backbone.
The network is implemented in PyTorch~\cite{pytorch} and optimized using Adam~\cite{kingma2015adam} with learning rate of 1e-3.
We train our model with batch size of 20, 40, and 20 for PASCAL-5$^{i}$, COCO-20$^{i}$, and FSS-1000 respectively.
We freeze the pre-trained backbone networks to prevent them from learning class-specific representations of the training data.
The intermediate tensor dimensions, the number of parameters of each layers and other additional details of the network are demonstrated in Tab.~\ref{tab:vgg16_arch}, \ref{tab:res50_arch}, and \ref{tab:res101_arch} for respective backbones of VGG16, ResNet50, and ResNet101.

\begin{table*}
    \begin{center}
    \scalebox{0.84}{
    \begin{tabular}{cl|cccccc|cccccc|c}
            \toprule
            \multirow{2}{*}{\shortstack{Backbone\\network}} & \multirow{2}{*}{Methods} & \multicolumn{6}{c|}{1-shot} & \multicolumn{6}{c|}{5-shot} & \# learnable \\ 
            
            & & $5^{0}$ & $5^{1}$ & $5^{2}$ & $5^{3}$ & mIoU & FB-IoU & $5^{0}$ & $5^{1}$ & $5^{2}$ & $5^{3}$ & mIoU & FB-IoU & params \\
            \midrule
            
            \multirow{3}{*}{VGG16~\cite{simonyan2015vgg}} & SG-One~\cite{zhang2020sgone}      & \underline{40.2} & \underline{58.4} & \underline{48.4} & \underline{38.4} & \underline{46.3} & 63.1 & \underline{41.9} & \underline{58.6} & \underline{48.6} & \underline{39.4} & 47.1 & 65.9 & \underline{19.0M} \\  
            
            & CRNet~\cite{liu2020crnet}    & - & - & - & - & \textbf{55.2} & \underline{66.4} & - & - & - & - & \underline{58.5} & \underline{71.0} & - \\   \cline{2-15} \\[-2.0ex]
            
            & HSNet (ours)          & \textbf{53.6} & \textbf{61.7} & \textbf{55.0} & \textbf{50.3} & \textbf{55.2} & \textbf{70.5} & \textbf{58.3} & \textbf{64.7} & \textbf{58.9} & \textbf{54.6} & \textbf{59.1} & \textbf{73.5} & \textbf{2.6M} \\
            
            \midrule
            
            \multirow{4}{*}{ResNet50~\cite{he2016deep}} & CANet~\cite{zhang2019canet}       & 52.5 & 65.9 & 51.3 & \underline{51.9} & 55.4 & 66.2 & 55.5 & \underline{67.8} & 51.9 & \underline{53.2} & 57.1 & 69.6 & \underline{19.0M} \\  
            
            & RPMM~\cite{yang2020pmm}        & \underline{55.2} & \textbf{66.9} & \underline{52.6} & 50.7 & \underline{56.3} & - & \underline{56.3} & 67.3 & \underline{54.5} & 51.0 & 57.3 & - & 19.7M \\  
            
            & CRNet~\cite{liu2020crnet}    & - & - & - & - & 55.7 & \underline{66.8} & - & - & - & - & \underline{58.8} & \underline{71.5} & - \\   \cline{2-15} \\[-2.0ex]
            
            & HSNet (ours)          & \textbf{57.4} & \underline{66.8} & \textbf{55.8} & \textbf{56.5} & \textbf{59.1} & \textbf{73.8} & \textbf{62.6} & \textbf{69.2} & \textbf{62.5} & \textbf{62.4} & \textbf{64.2} & \textbf{77.4} & \textbf{2.6M} \\
            
            \midrule
            \multirow{1}{*}{ResNet101~\cite{he2016deep}} & HSNet (ours)          & \textbf{60.1} & \textbf{67.8} & \textbf{57.3} & \textbf{59.0} & \textbf{61.1} & \textbf{74.4} & \textbf{63.9} & \textbf{69.9} & \textbf{62.0} & \textbf{63.6} & \textbf{64.8} & \textbf{77.1} & \textbf{2.6M} \\
            
            \bottomrule
    \end{tabular}
    }
    \vspace{0.0mm}
    \caption{\label{tab:pascal_sota_noignore}Evaluation results on PASCAL-5$^{i}$~\cite{shaban2017oslsm} benchmark in mIoU and FB-IoU evaluation metrics without the use of \texttt{ignore\_label}. The results of other methods are from~\cite{liu2020crnet,liu2020ppnet,tian2020pfenet,wang2020dan,yang2020pmm}.}
    \vspace{-6.0mm}
    \end{center}
\end{table*}

\section{Additional results and analyses}

\smallbreak
\noindent \textbf{Additional $K$-shot results.}
Following the work of~\cite{malik2020repri,tian2020pfenet,yang2020pmm}, we conduct $K$-shot experiments with $K \in \{1, 5, 10\}$.
Table~\ref{tab:1_5_10-shot} compares our results with the recent methods~\cite{malik2020repri,tian2020pfenet,yang2020pmm} on PASCAL-5$^{i}$ and COCO-20$^{i}$.
The significant performance improvements on both datasets clearly indicate the effectiveness of our approach.
Achieving 2.5\%p and 4.6\%p mIoU improvements over the previous best method~\cite{malik2020repri} on respective PASCAL-5$^{i}$ and COCO-20$^{i}$, our model again sets a new state of the art in 10-shot setting as well, showing notable improvements with larger $K$.

\smallbreak
\noindent \textbf{Numerical comparisons of ablation study.}
We tabularize Figures \ref{fig:hypercorr_ablation_quant} and \ref{fig:pyramid_ablation_quant}, \eg, ablation study on hypercorrelations and pyramidal layers, in Tables~\ref{tab:ablation_hypercorr} and \ref{tab:ablation_pyramid} respectively.
Achieving 4.5\%p mIoU improvements over $\mathbf{C}_{p}^{\text{deep}}$, our method clearly benefits from diverse feature correlations from multi-level CNN layers ($\mathbf{C}_{p}$) as seen in Tab.~\ref{tab:ablation_hypercorr}.
A large performance gap between $\mathcal{C}^{(2:3)}$ and $\mathcal{C}^{(3)}$ in Tab.~\ref{tab:ablation_pyramid} (63.9 vs. 55.5) reveals that the intermediary second pyramidal layer ($p=2$) is especially effective in robust mask prediction compared to the first pyramidal layer ($p=1$).

\begin{table}[t]

    \begin{center}
        \scalebox{0.8}{
        \begin{tabular}{l|ccc|ccc}
                \toprule
                \multirow{2}{*}{Method} & \multicolumn{3}{c|}{PASCAL-5$^{i}$} & \multicolumn{3}{c}{COCO-20$^{i}$} \\
                
                & 1-shot & 5-shot & 10-shot & 1-shot & 5-shot & 10-shot \\
                
                \midrule
                
                RPMM~\cite{yang2020pmm} & 56.3 & 57.3 & 57.6 & 30.6 & 35.5 & 33.1 \\  
                
                PFENet~\cite{tian2020pfenet}  & \underline{60.8} & 61.9 & 62.1 & \underline{35.8} & 39.0 & 39.7 \\
                
                RePRI~\cite{malik2020repri}  & 59.7 & \underline{66.6} & \underline{68.1} & 34.1 & \underline{41.6} & \underline{44.1} \\   \cline{1-7} \\[-2.0ex]
                
                HSNet (ours)  & \textbf{64.0} & \textbf{69.5} & \textbf{70.6} & \textbf{39.2} & \textbf{46.9} & \textbf{48.7} \\
                
                \bottomrule
        \end{tabular}
        }
    \caption{\label{tab:1_5_10-shot}Results on $K$-shot with ResNet50 backbone network where $K \in \{1, 5, 10\}$. The results of other methods are from~\cite{malik2020repri}.}
    \vspace{-5.0mm}
    \end{center}
\end{table}

\begin{table}[t]
    \begin{center}
    \scalebox{0.65}{
    \begin{tabular}{l|ccccc|ccccc}
            \toprule
            \multirow{2}{*}{Methods} & \multicolumn{5}{c|}{1-shot} & \multicolumn{5}{c}{5-shot} \\ 
            
            & $5^{0}$ & $5^{1}$ & $5^{2}$ & $5^{3}$ & mean & $5^{0}$ & $5^{1}$ & $5^{2}$ & $5^{3}$ & mean \\ \\[-2.0ex]
            
            \midrule
            
            $\mathbf{C}_{p}^{\text{shallow}}$ & 57.1 & 64.7 & 57.5 & 57.0 & 59.1 & 63.7 & 69.8 & \underline{66.0} & 63.4 & 65.7 \\  \\[-2.0ex]
            
            $\mathbf{C}_{p}^{\text{deep}}$ & \underline{60.6} & \underline{68.6} & \underline{58.2} & \underline{59.2} & \underline{61.7} & \underline{66.7} & \underline{72.0} & 65.9 & \underline{65.4} & \underline{67.5} \\ \cline{1-11} \\[-2.0ex]
            
            $\mathbf{C}_{p}$ (ours)    & \textbf{67.3} & \textbf{72.3} & \textbf{62.0} & \textbf{63.1} & \textbf{66.2} & \textbf{71.8} & \textbf{74.4} & \textbf{67.0} & \textbf{68.3} & \textbf{70.4} \\
            
            \bottomrule
    \end{tabular}
    }
    \caption{\label{tab:ablation_hypercorr}Numerical results of Figure \ref{fig:hypercorr_ablation_quant}. All experiments are performed with ResNet101 backbone~\cite{he2016deep}.}
    \vspace{-5.0mm}
    \end{center}
\end{table}

\begin{table}[t]
    \begin{center}
    \scalebox{0.65}{
    \begin{tabular}{l|ccccc|ccccc}
            \toprule
            \multirow{2}{*}{Methods} & \multicolumn{5}{c|}{1-shot} & \multicolumn{5}{c}{5-shot}  \\ 
            
            & $5^{0}$ & $5^{1}$ & $5^{2}$ & $5^{3}$ & mean & $5^{0}$ & $5^{1}$ & $5^{2}$ & $5^{3}$ & mean \\
            
            \midrule
            
            $\mathcal{C}^{(3)}$ & 54.3 & 62.8 & 52.0 & 52.8 & 55.5 & 60.2 & 67.0 & 59.4 & 59.9 & 61.6 \\  
            
            $\mathcal{C}^{(2:3)}$ & \underline{64.3} & \underline{70.3} & \underline{60.5} & \underline{60.4} & \underline{63.9} & \underline{69.7} & \underline{73.2} & \underline{65.2} & \underline{64.9} & \underline{68.2} \\ \cline{1-11} \\[-2.0ex]
            
            $\mathcal{C}$ (ours)    & \textbf{67.3} & \textbf{72.3} & \textbf{62.0} & \textbf{63.1} & \textbf{66.2} & \textbf{71.8} & \textbf{74.4} & \textbf{67.0} & \textbf{68.3} & \textbf{70.4} \\
            
            \bottomrule
    \end{tabular}
    }
    \caption{\label{tab:ablation_pyramid}Numerical results of Figure \ref{fig:pyramid_ablation_quant}. All experiments are performed with ResNet101 backbone~\cite{he2016deep}.}
    \vspace{-5.0mm}
    \end{center}
\end{table}

\smallbreak
\noindent \textbf{Evaluation results without using} \texttt{ignore\_label} \textbf{on PASCAL-5$^{i}$.}
The benchmarks of PASCAL-5$^{i}$~\cite{shaban2017oslsm}, COCO-20$^{i}$~\cite{lin2015coco}, and FSS-1000~\cite{li2020fss1000} consist of segmentation mask annotations in which each pixel is labeled with either background or one of the predefined object categories.
As pixel-wise segmentation near object boundaries is ambiguous to perform even for human annotators, PASCAL-5$^{i}$ uses a special kind of label called \texttt{ignore\_label} which marks pixel regions ignored during training and evaluation to mitigate the ambiguity\footnote{The use of \texttt{ignore\_label} was originally adopted in PASCAL VOC dataset~\cite{everingham2015pascal}. The same evaluation criteria is naturally transferred to PASCAL-5$^{i}$~\cite{shaban2017oslsm} as it is created from PASCAL VOC.}.

Most recent few-shot segmentation work~\cite{malik2020repri,liu2020ppnet,nguyen2019fwb,rakelly2018cofcn,shaban2017oslsm,siam2019amp,tian2020pfenet,wang2020dan,wang2019panet,zhang2019pgnet} adopt this evaluation criteria but we found that some methods~\cite{liu2020crnet,yang2020pmm,zhang2019canet,zhang2020sgone} do not utilize \texttt{ignore\_label} in their evaluations.
Therefore, the methods are unfairly evaluated as fine-grained mask prediction near object boundaries is one of the most challenging part in segmentation problem.
For fair comparisons, we intentionally exclude the methods of~\cite{liu2020crnet,yang2020pmm,zhang2019canet,zhang2020sgone} from Tab.~\ref{tab:pascal_sota} and compare the results of our model evaluated without the use of \texttt{ignore\_label} with those methods~\cite{liu2020crnet,yang2020pmm,zhang2019canet,zhang2020sgone}.
The results are summarized in Tab.~\ref{tab:pascal_sota_noignore}.
Even without using \texttt{ignore\_label}, the proposed method sets a new state of the art with ResNet50 backbone, outperforming the previous best methods of \cite{yang2020pmm} and \cite{liu2020crnet} by (1-shot) 2.8\%p and (5-shot) 5.4\%p respectively.
With VGG16 backbone, our method performs comparably effective to the previous best method~\cite{liu2020crnet} while having the smallest learnable parameters.

\section{Qualitative results}

\begin{figure}[t]
    \begin{center}
        \includegraphics[width=0.95\linewidth]{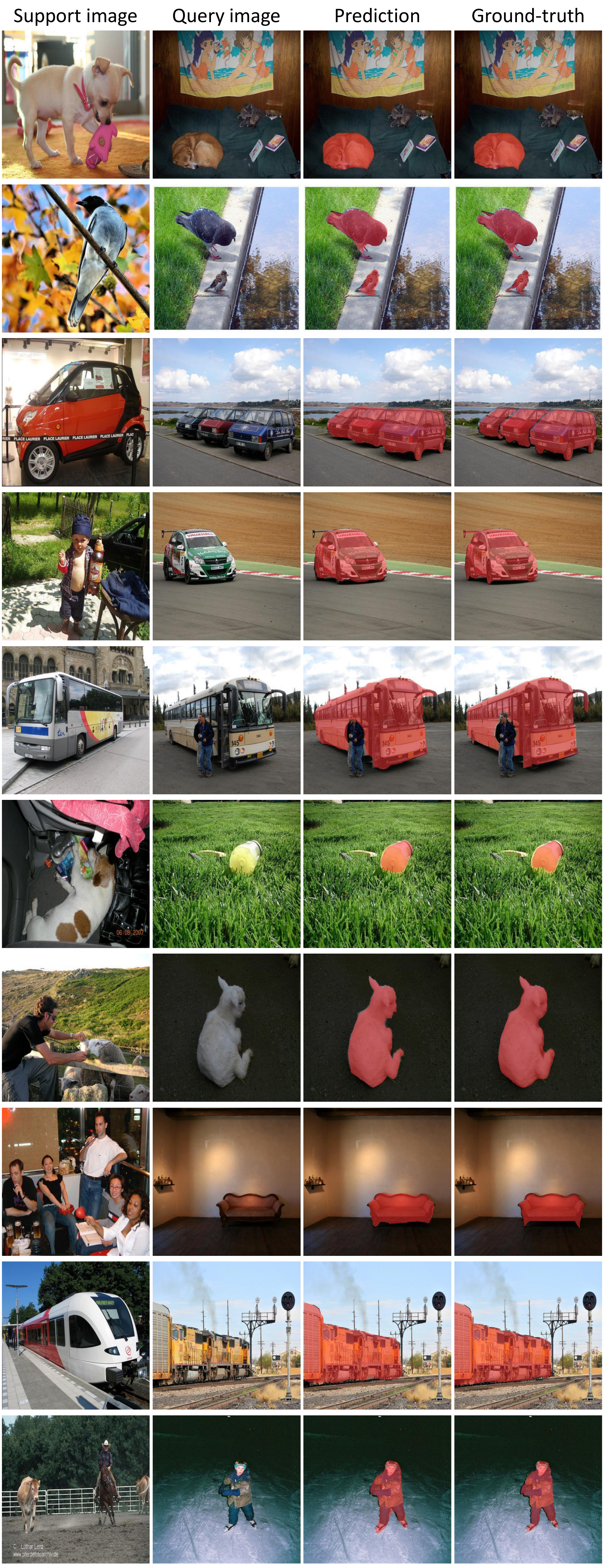}
    \end{center}
    \vspace{-6.0mm} 
        \caption{Example results without support feature masking (Eqn.~\ref{eqn:masking}) on PASCAL-5$^{i}$ dataset.}
    \vspace{-3.0mm} 
    \label{fig:qual_without_masking}
\end{figure}

\smallbreak
\noindent \textbf{Results without support feature masking.}
As demonstrated in Sec.~\ref{sec:results_analyses}, we conduct experiments without support feature masking (Eqn.~\ref{eqn:masking}), similarly to co-segmentation problem with stronger demands for generalizibility.
Figure~\ref{fig:qual_without_masking} visualizes some example results on PASCAL-5$^{i}$ dataset.
Even without the use of support masks (in both training and testing), our model effectively segments target instances in query images.
The results indicate that learning patterns of feature correlations from multiple visual aspects is effective in fine-grained segmentation as well as identifying `common' instances in the support and query images.

\begin{figure}[t]
    \begin{center}
        \includegraphics[width=0.95\linewidth]{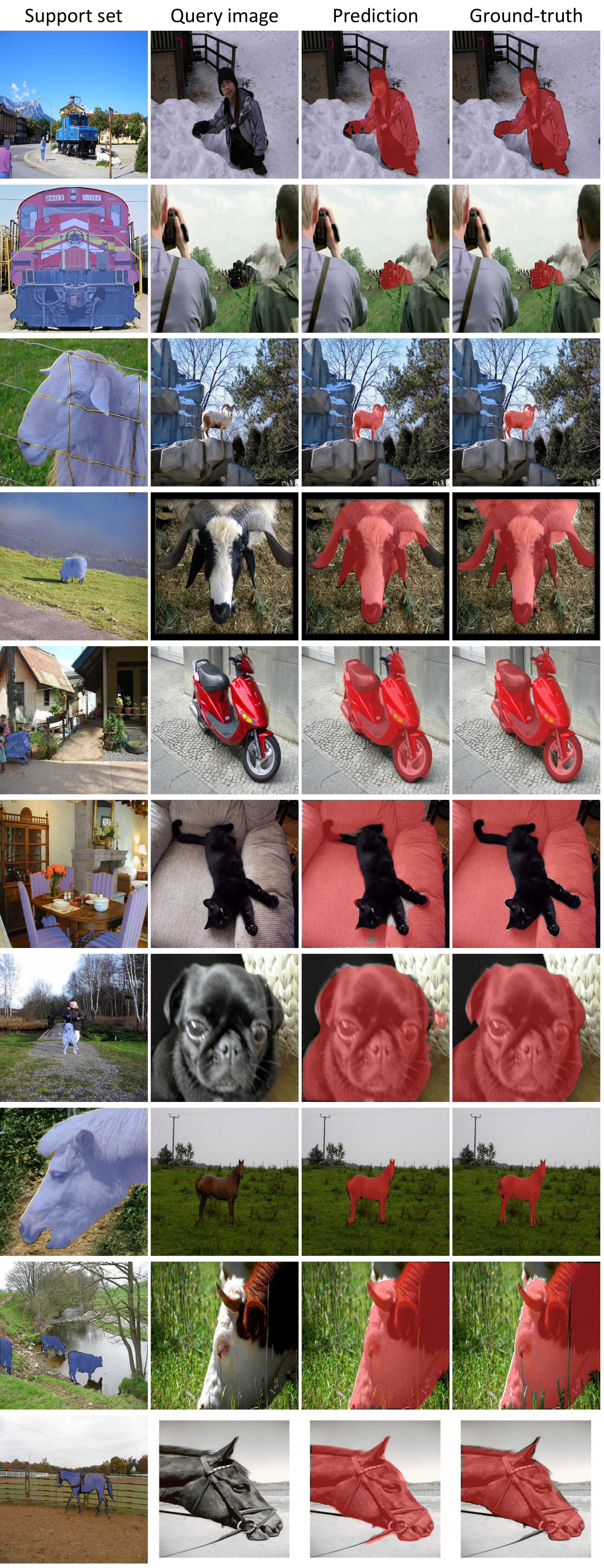}
    \end{center}
    \vspace{-5.0mm} 
        \caption{Qualitative (1-shot) results on PASCAL-5$^{i}$~\cite{shaban2017oslsm} dataset under large differences in object scales.}
    \vspace{-3.0mm} 
    \label{fig:qual_scale_diff_pas}
\end{figure}

\begin{figure}[t]
    \begin{center}
        \includegraphics[width=0.95\linewidth]{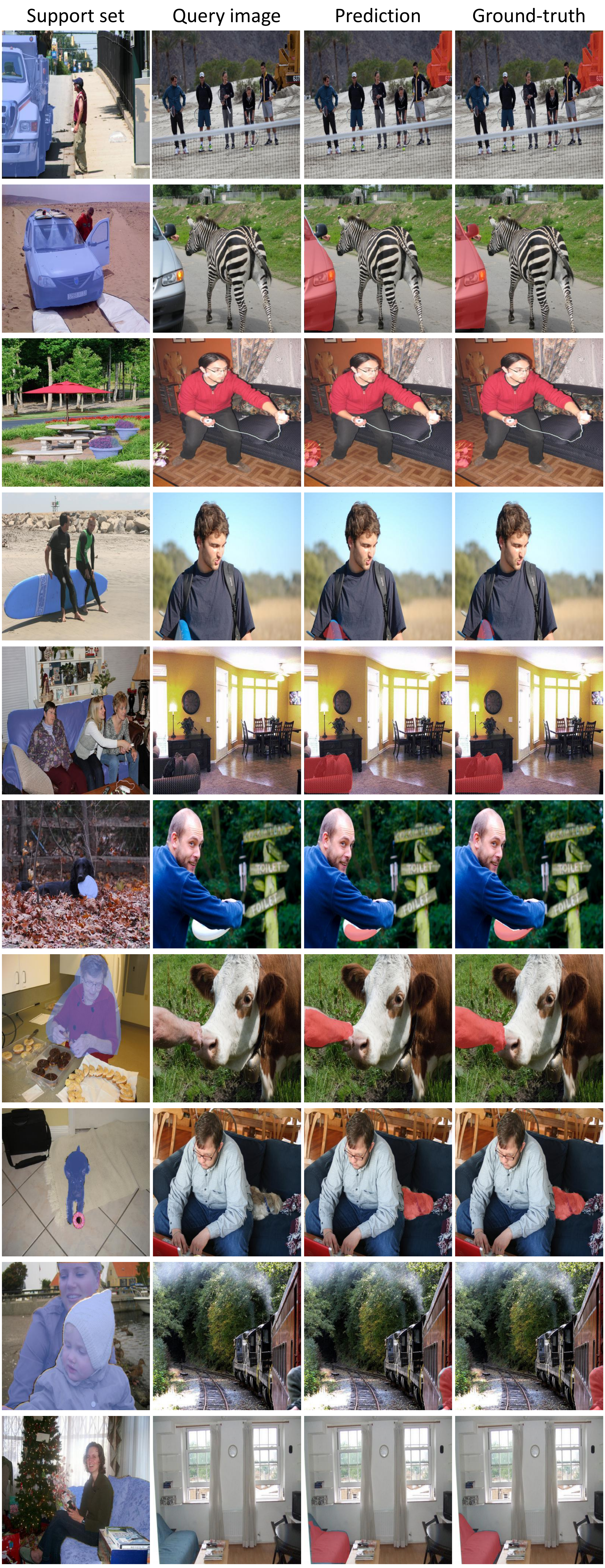}
    \end{center}
    \vspace{-5.0mm} 
        \caption{Qualitative (1-shot) results on PASCAL-5$^{i}$~\cite{shaban2017oslsm} and COCO-20$^{i}$~\cite{lin2015coco} datasets under large truncations and occlusions.}
    \vspace{-3.0mm} 
    \label{fig:qual_trunc_occ}
\end{figure}

\begin{figure}[t]
    \begin{center}
        \includegraphics[width=0.95\linewidth]{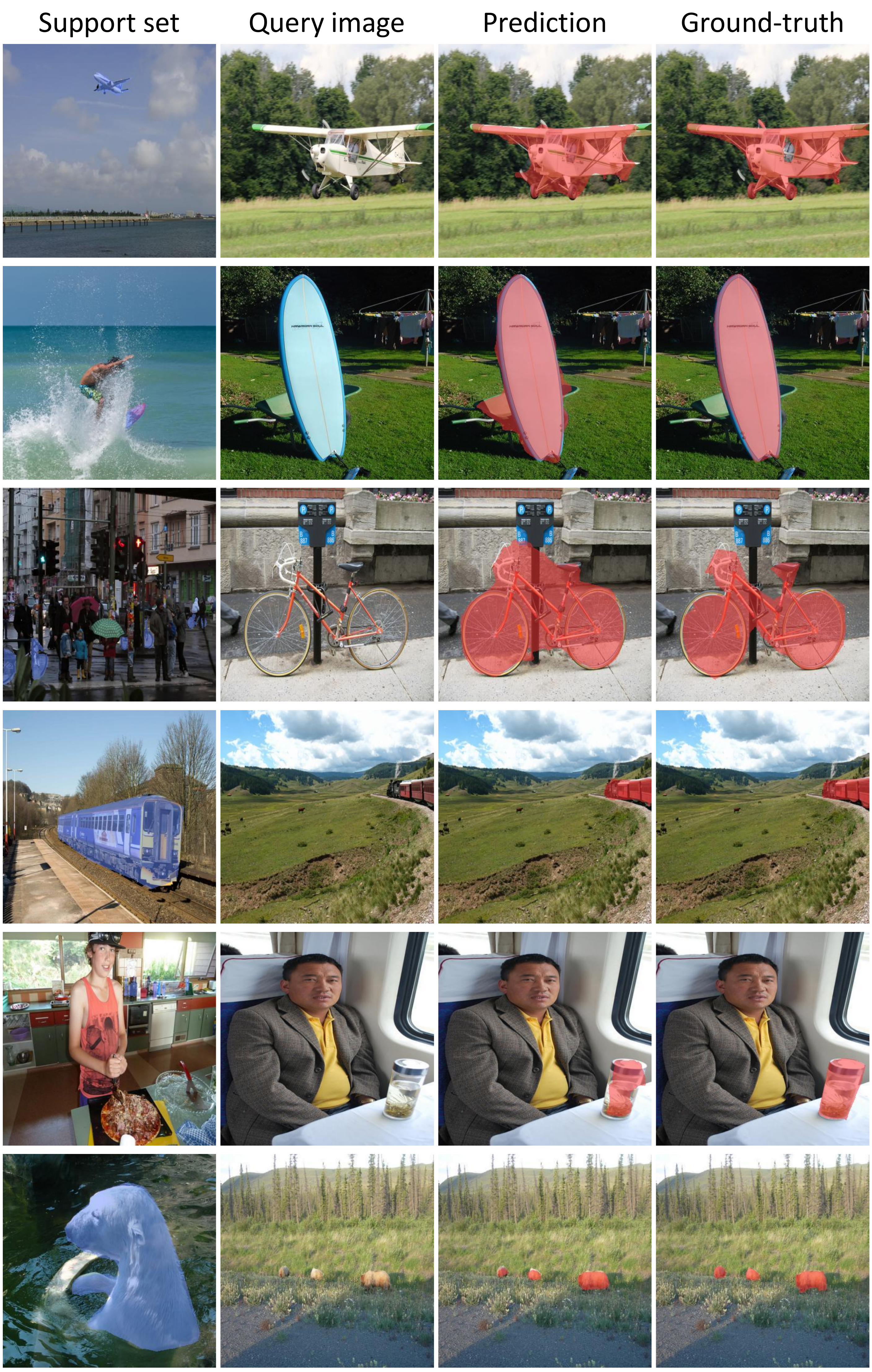}
    \end{center}
    \vspace{-5.0mm} 
        \caption{Qualitative (1-shot) results on COCO-20$^{i}$~\cite{lin2015coco} dataset under large differences in object scales.}
    \vspace{-2.0mm} 
    \label{fig:qual_scale_diff_coco}
\end{figure}

\begin{figure}[t]
    \begin{center}
        \includegraphics[width=0.95\linewidth]{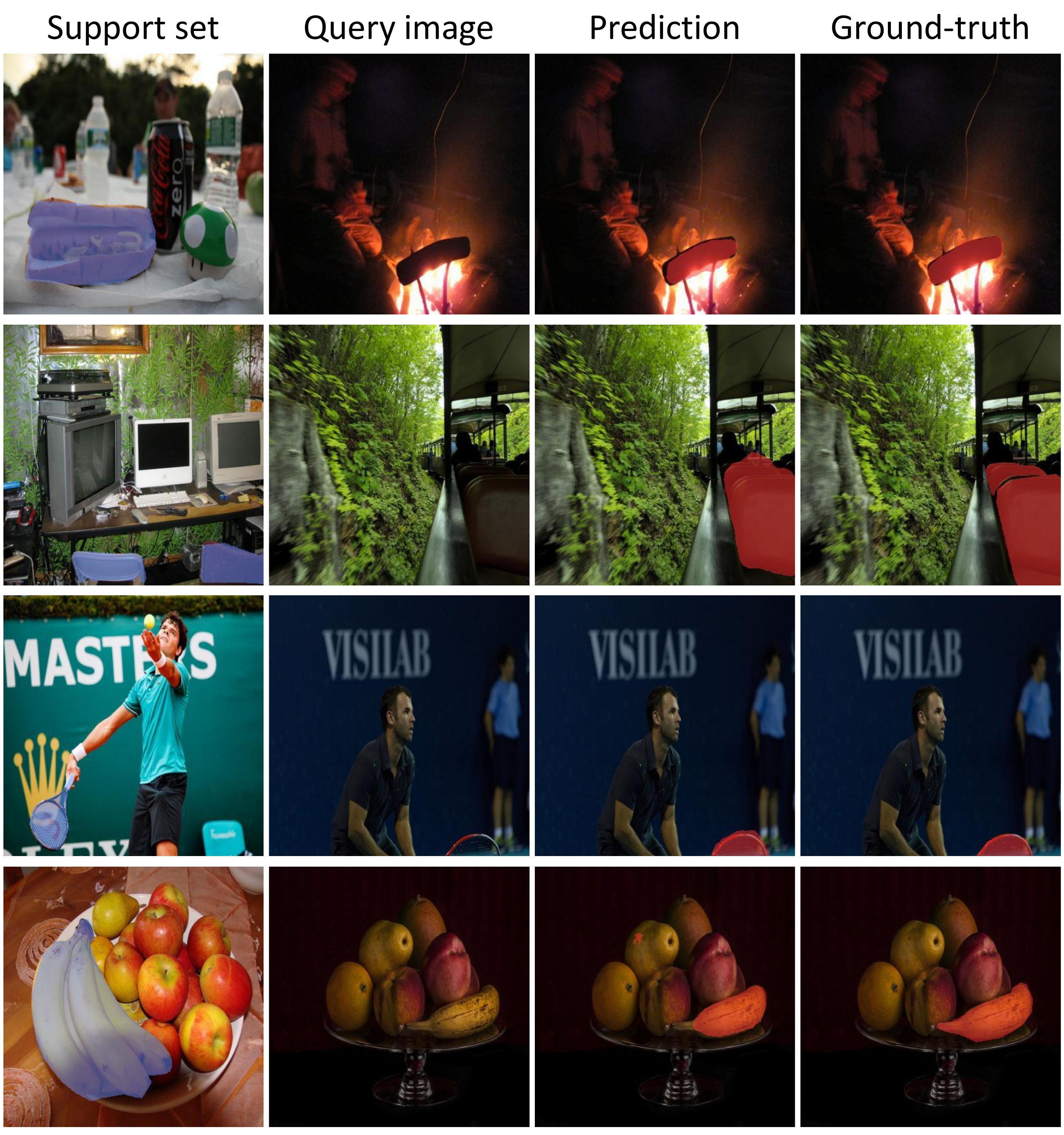}
    \end{center}
    \vspace{-6.0mm} 
        \caption{Qualitative (1-shot) results on COCO-20$^{i}$~\cite{lin2015coco} dataset under large illumination-changes in support and query images.}
    \vspace{-3.0mm} 
    \label{fig:qual_illumination}
\end{figure}

\begin{figure}[t]
    \begin{center}
        \includegraphics[width=0.95\linewidth]{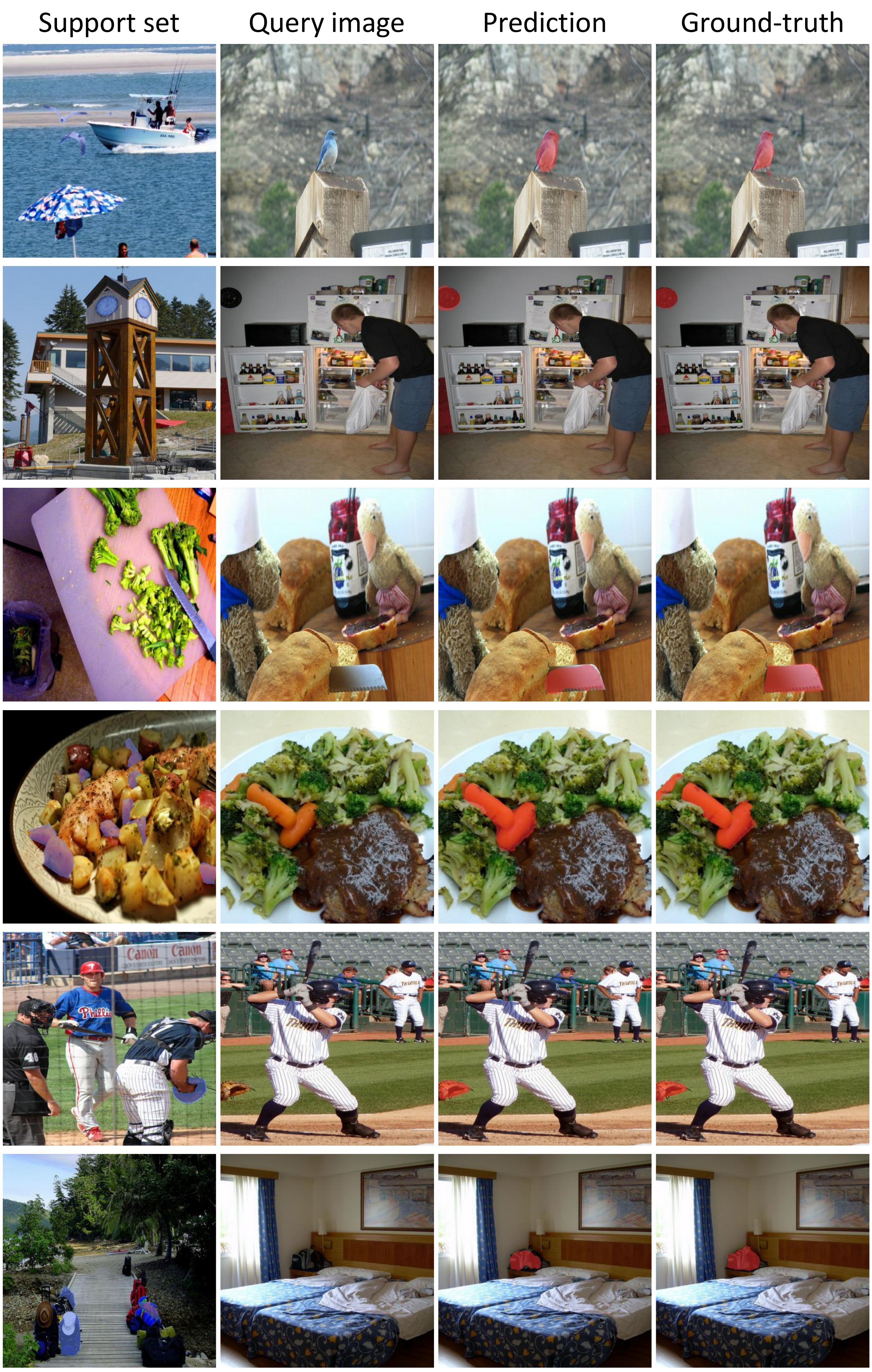}
    \end{center}
    \vspace{-6.0mm} 
        \caption{Qualitative (1-shot) results on COCO-20$^{i}$~\cite{lin2015coco} dataset with exceptionally small objects.}
    \vspace{-2.0mm} 
    \label{fig:qual_small_obj}
\end{figure}

\begin{figure}[t]
    \begin{center}
        \includegraphics[width=0.95\linewidth]{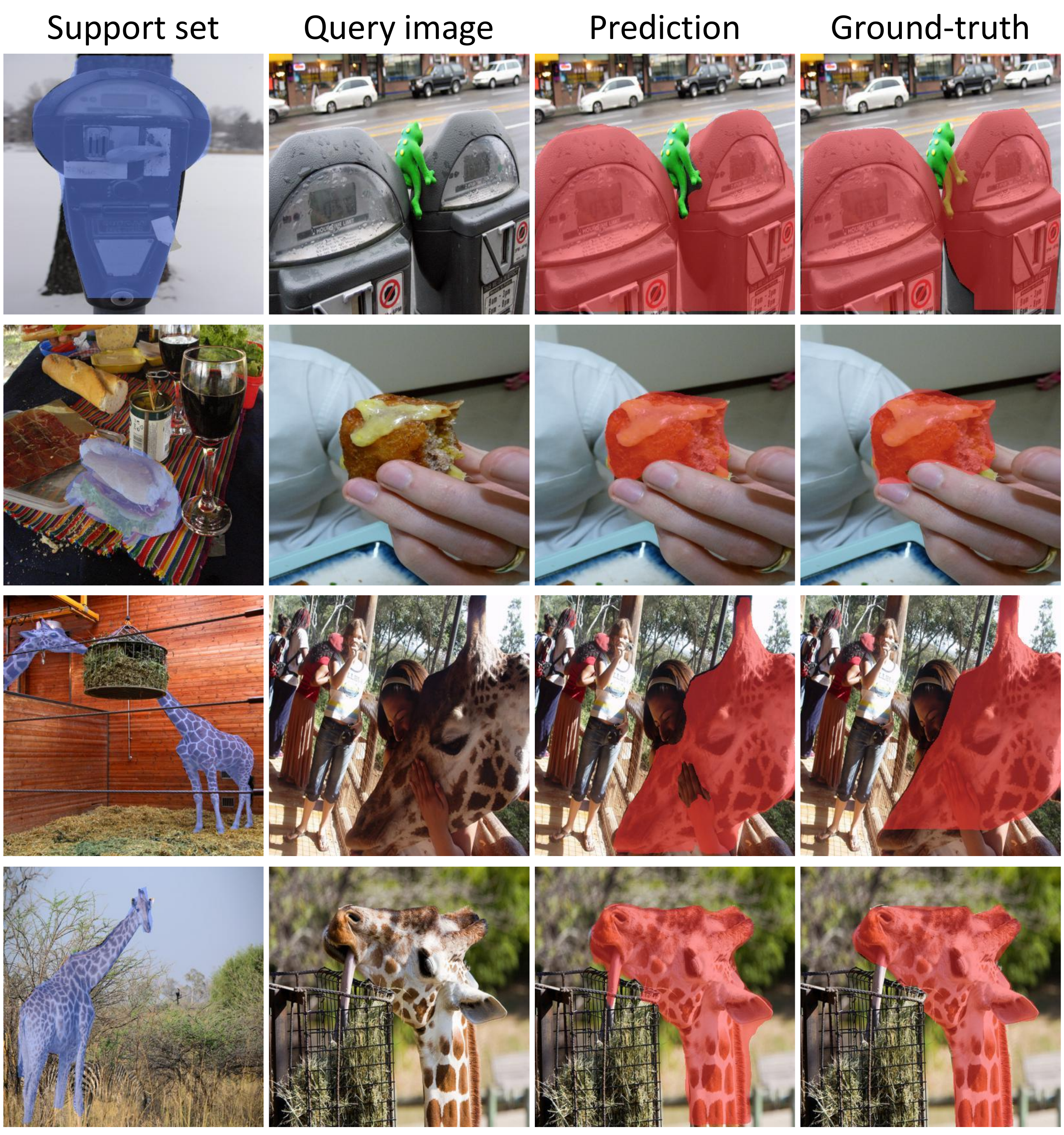}
    \end{center}
    \vspace{-6.0mm} 
        \caption{Our network occasionally predicts more accurate segmentation masks than human-annotated ground-truths.}
    \vspace{-3.0mm} 
    \label{fig:qual_more_accurate}
\end{figure}

\begin{figure}[t]
    \begin{center}
        \includegraphics[width=0.95\linewidth]{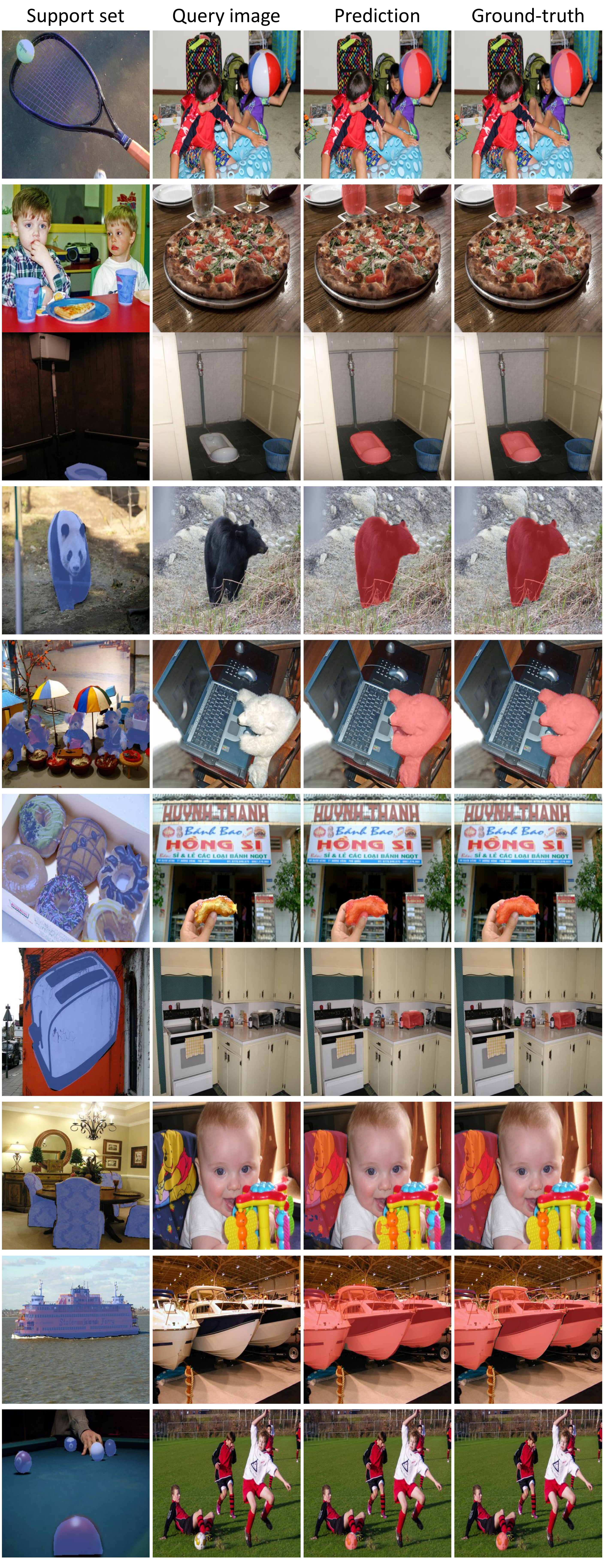}
    \end{center}
    \vspace{-6.0mm} 
        \caption{Qualitative (1-shot) results on PASCAL-5$^{i}$~\cite{shaban2017oslsm} and COCO-20$^{i}$~\cite{lin2015coco} datasets under large intra-class variations.}
    \vspace{-3.0mm} 
    \label{fig:qual_intra_class}
\end{figure}

\begin{figure}[t]
    \begin{center}
        \includegraphics[width=0.95\linewidth]{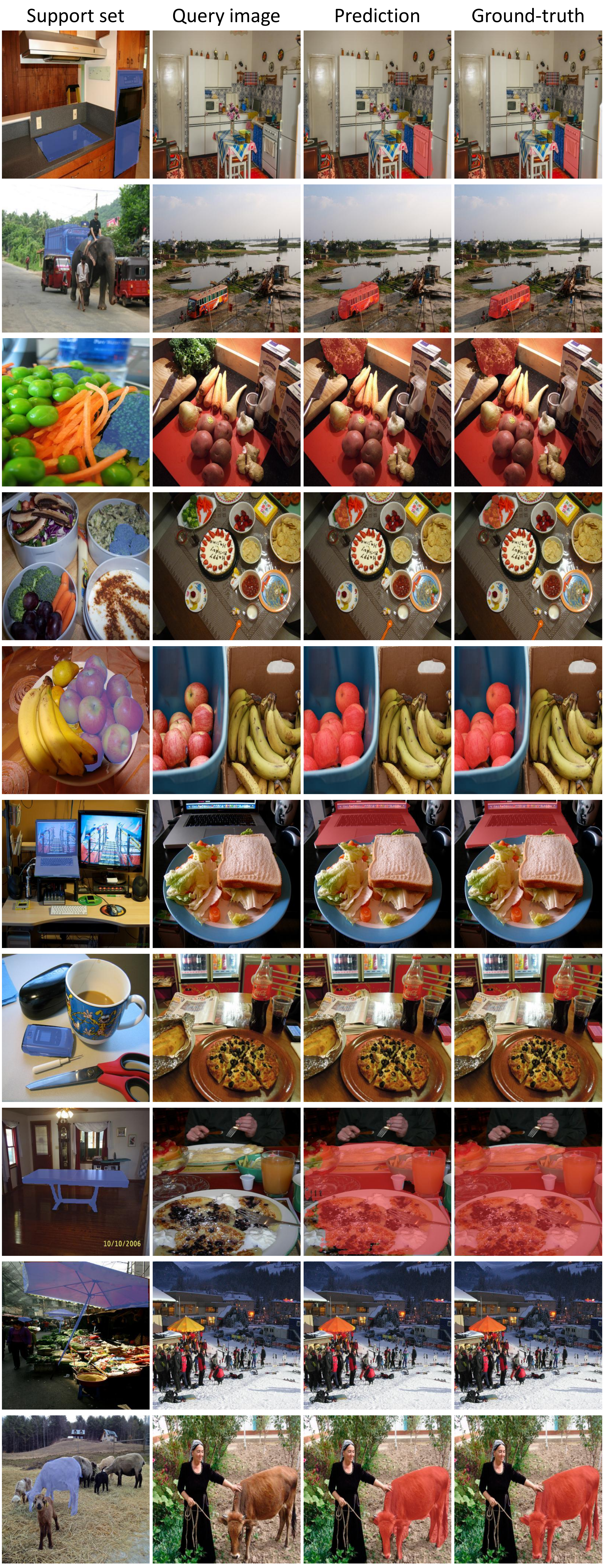}
    \end{center}
    \vspace{-6.0mm} 
        \caption{Qualitative (1-shot) results on PASCAL-5$^{i}$~\cite{shaban2017oslsm} and COCO-20$^{i}$~\cite{lin2015coco} datasets in presence of noisy background clutters.}
    \vspace{-3.0mm} 
    \label{fig:qual_clutters}
\end{figure}

\begin{figure}[t]
    \begin{center}
        \includegraphics[width=0.95\linewidth]{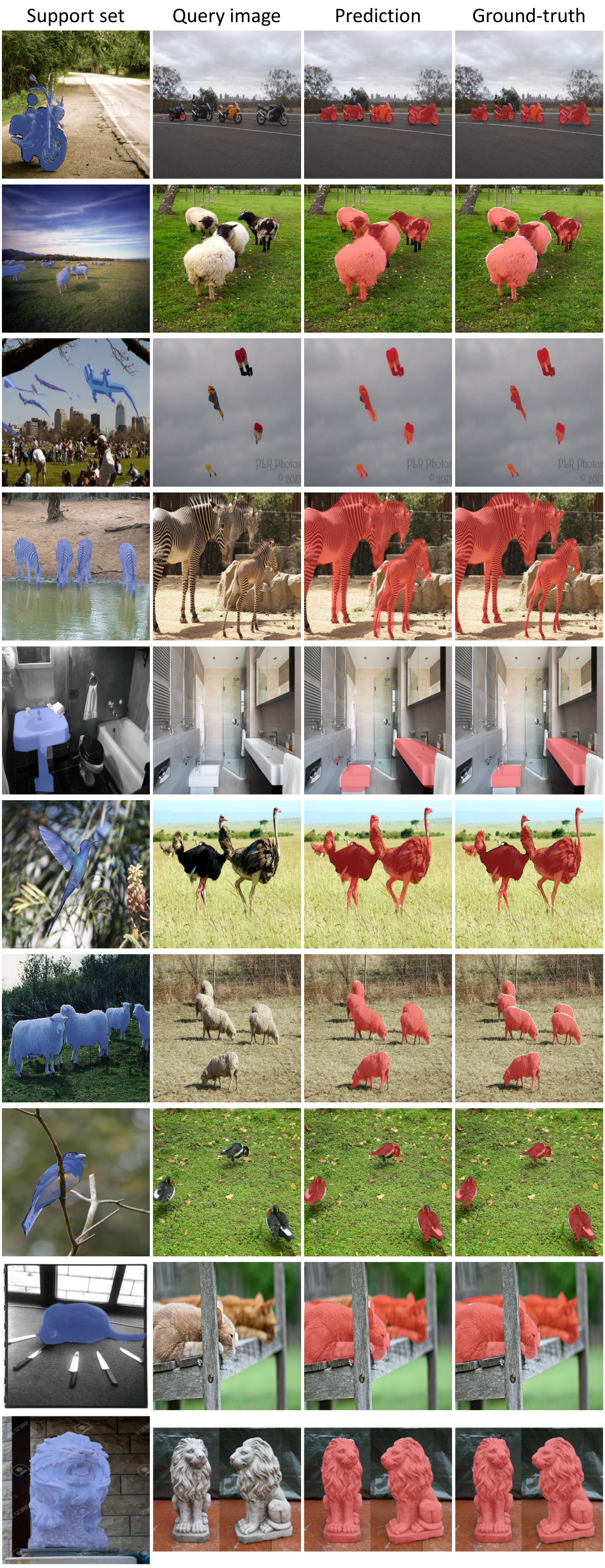}
    \end{center}
    \vspace{-6.0mm} 
        \caption{One-to-many and many-to-many (1-shot) results on PASCAL-5$^{i}$~\cite{shaban2017oslsm}, COCO-20$^{i}$~\cite{lin2015coco}, and FSS-1000~\cite{li2020fss1000} datasets.}
    \vspace{-3.0mm} 
    \label{fig:qual_manytomany}
\end{figure}

\begin{figure}[t]
    \begin{center}
        \includegraphics[width=0.95\linewidth]{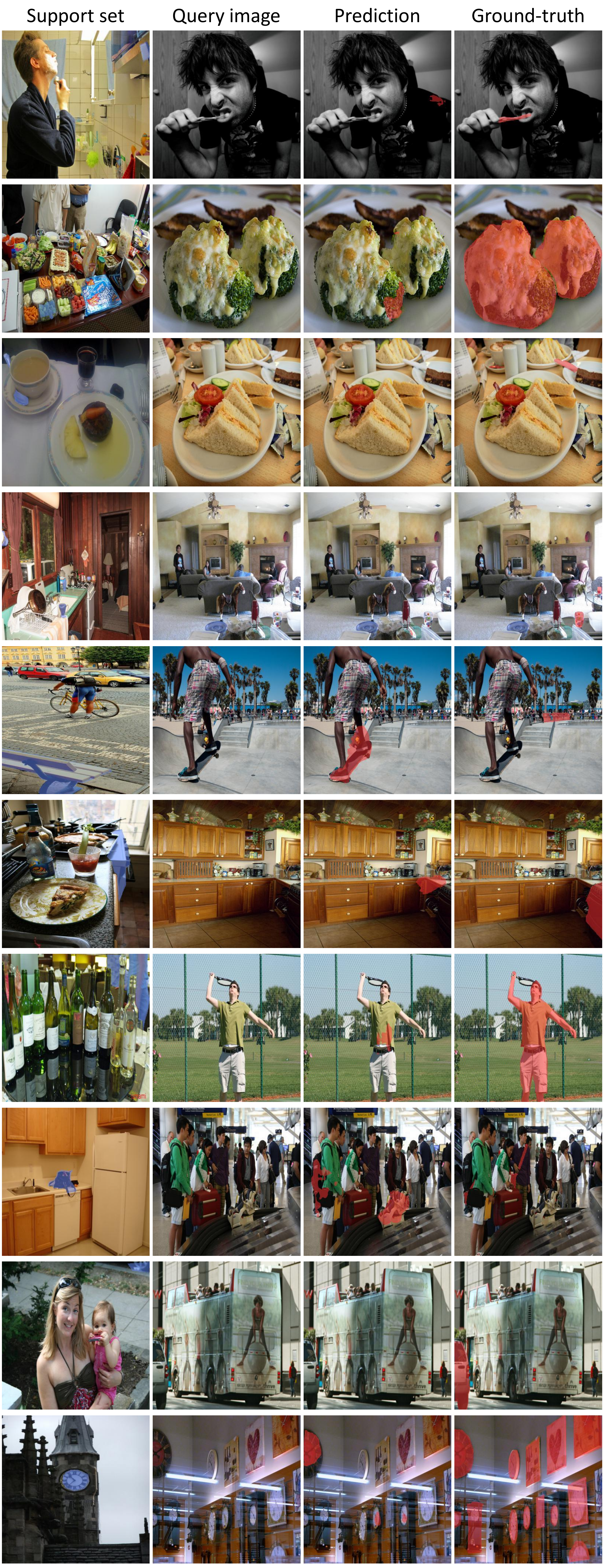}
    \end{center}
    \vspace{-6.0mm} 
        \caption{Representative failure cases in presence of severe occlusions, intra-class variances and extremely tiny objects.}
    \vspace{-3.0mm} 
    \label{fig:qual_failure}
\end{figure}

\begin{figure*}
    \begin{center}
        \includegraphics[width=0.81\linewidth]{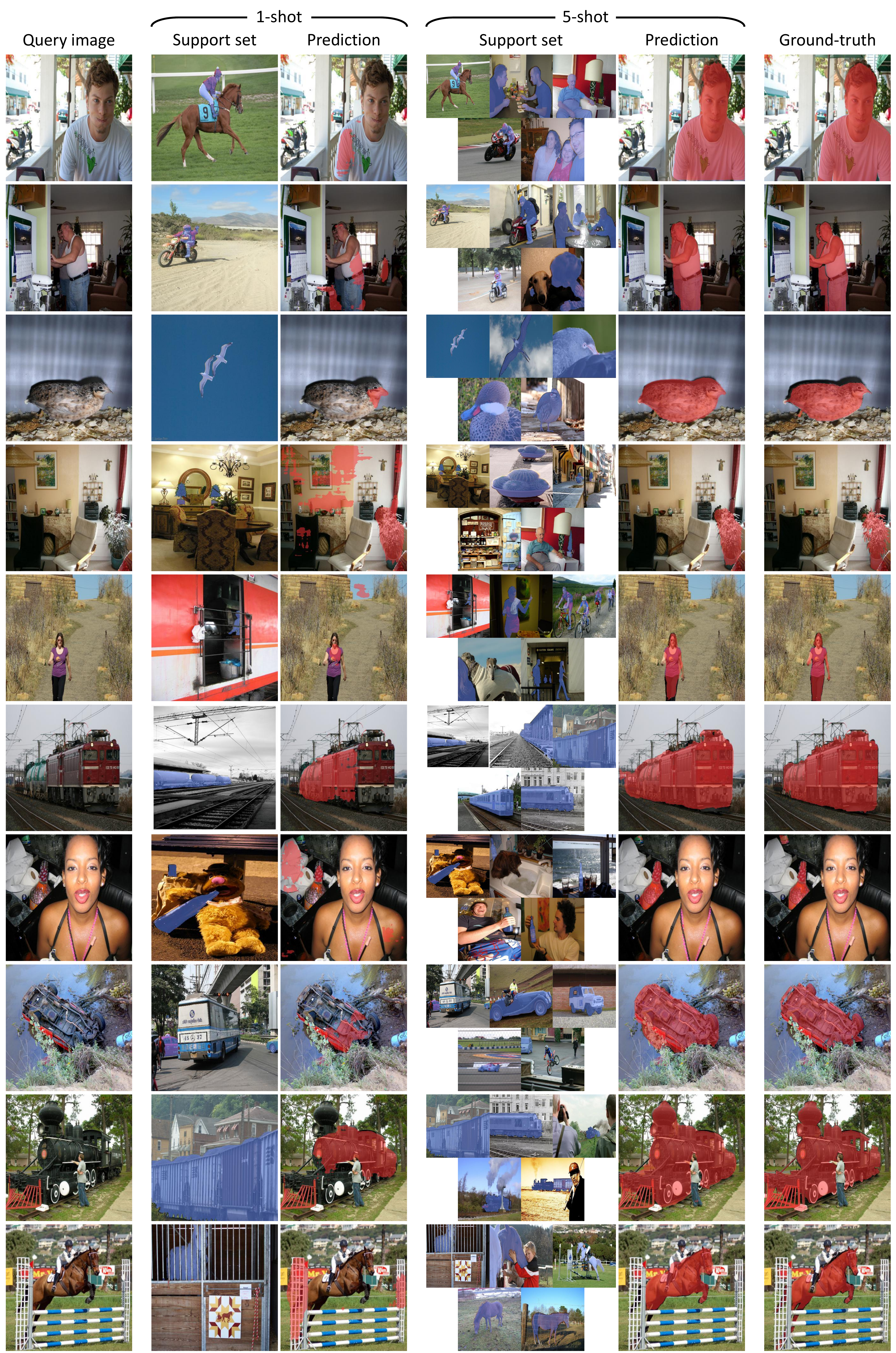}
    \end{center}
    \vspace{-6.0mm} 
        \caption{Comparison between 1-shot and 5-shot results on PASCAL-5$^{i}$ dataset~\cite{shaban2017oslsm}. Multiple support images and mask annotations clearly help our model generate accurate mask predictions on query images in many challenging cases.}
    \vspace{-3.0mm} 
    \label{fig:qual_kshot}
\end{figure*}

\smallbreak
\noindent \textbf{Additional qualitative results.}
We present additional qualitative results on PASCAL-5$^{i}$~\cite{shaban2017oslsm}, COCO-20$^{i}$~\cite{nguyen2019fwb}, and FSS-1000~\cite{li2020fss1000} benchmark datasets.
All the qualitative results are best viewed in electronic forms.
Example results in presence of large scale-differences, truncations, and occlusions are shown in Fig.~\ref{fig:qual_scale_diff_pas}, \ref{fig:qual_trunc_occ}, and \ref{fig:qual_scale_diff_coco}.
Figure~\ref{fig:qual_illumination} visualizes model predictions under large illumination-changes in support and query images.
Figure~\ref{fig:qual_small_obj} visualizes some sample predictions given exceptionally small objects in either support or query images.
As seen in Fig.~\ref{fig:qual_more_accurate}, we found that our model sometimes predicts more reliable segmentation masks than ground-truth ones.
Some qualitative results in presence of large intra-class variations and noisy clutters in background are shown in Fig.~\ref{fig:qual_intra_class} and \ref{fig:qual_clutters}.
Given only a single support image-annotation pair, our model effectively segments multiple instances in a query image as visualized in Fig.~\ref{fig:qual_manytomany}. 
Figure~\ref{fig:qual_failure} shows representative failure cases;
our model fails to localize target objects in presence of severe occlusions, intra-class variances and extremely tiny support (or query) objects.
As seen in Fig.~\ref{fig:qual_kshot}, the model predictions become much reliable given multiple support image-mask pairs, \ie, $K>1$.

The code and data to reproduce all experiments in this paper is available at our project page: \url{http://cvlab.postech.ac.kr/research/HSNet/}.

\clearpage

\clearpage

\clearpage

\clearpage

\clearpage

\clearpage

\begin{table*}
    \begin{center}
    \scalebox{0.88}{
    \begin{tabular}{lcccccc}
            \toprule
            \textbf{Layer} & \multicolumn{2}{c}{\textbf{Input}} & \multicolumn{2}{c}{\textbf{Output}} & \textbf{Operation} & \textbf{\# params.} \\
            
            \midrule
            
            \multirow{6}{*}{VGG16 Backbone} & \multirow{3}{*}{$I^{\mathrm{q}}$} & \multirow{3}{*}{(3, 400, 400)} & \multirow{3}{*}{$\{\mathbf{F}_{l}^{\mathrm{q}}\}_{l=1}^{7}$} & (512, 12, 12) $\times$ 1 & \multirow{6}{*}{\textsc{Series of 2D Convs}} & \multirow{6}{*}{\shortstack{14.7M\\(frozen)}} \\
            
            & & & & (512, 25, 25) $\times$ 3 & & \\
            & & & & (512, 50, 50) $\times$ 3 & & \\ \cline{2-5} \\[-2.0ex]
            
            & \multirow{3}{*}{$I^{\mathrm{s}}$} & \multirow{3}{*}{(3, 400, 400)} & \multirow{3}{*}{$\{\mathbf{F}_{l}^{\mathrm{s}}\}_{l=1}^{7}$} & (512, 12, 12) $\times$ 1 &  &  \\
            & & & & (512, 25, 25) $\times$ 3 & & \\
            & & & & (512, 50, 50) $\times$ 3 & & \\
            
            \midrule
            
            \multirow{4}{*}{Masking Layer} & \multirow{3}{*}{$\{\mathbf{F}_{l}^{\mathrm{s}}\}_{l=1}^{7}$} & (512, 12, 12) $\times$ 1 & \multirow{4}{*}{$\{\hat{\mathbf{F}}_{l}^{\mathrm{s}}\}_{l=1}^{7}$} & \multirow{4}{*}{\shortstack{(512, 12, 12) $\times$ 1\\(512, 25, 25) $\times$ 3\\(512, 50, 50) $\times$ 3}} & \multirow{4}{*}{\shortstack{\textsc{Bilinear Interpolation}\\\textsc{Hadamard Product}}} & \multirow{4}{*}{-} \\
            
            &  & (512, 25, 25) $\times$ 3 &  &  &  &  \\
            &  & (512, 50, 50) $\times$ 3 &  &  &  &  \\ \cline{2-3} \\[-2.0ex]
            & $\mathbf{M}^{\mathrm{s}}$ & (1, 400, 400) &  &  &  &  \\
            
            \midrule
            
            \multirow{6}{*}{Correlation Layer} & \multirow{3}{*}{$\{\mathbf{F}_{l}^{\mathrm{q}}\}_{l=1}^{7}$} & (512, 12, 12) $\times$ 1 & \multirow{6}{*}{$\{\mathbf{C}_{p}\}_{p=1}^{3}$} & \multirow{6}{*}{\shortstack{(1, 12, 12, 12, 12)\\(3, 25, 25, 25, 25)\\(3, 50, 50, 50, 50)}} & \multirow{6}{*}{\textsc{Cosine Similarity}} & \multirow{6}{*}{-} \\
            
            &  & (512, 25, 25) $\times$ 3 &  &  &  &  \\
            &  & (512, 50, 50) $\times$ 3 &  &  &  &  \\ \cline{2-3} \\[-2.0ex]
            & \multirow{3}{*}{$\{\hat{\mathbf{F}}_{l}^{\mathrm{s}}\}_{l=1}^{7}$} & (512, 12, 12) $\times$ 1 &  &  &  &  \\
            &  & (512, 25, 25) $\times$ 3 &  &  &  &  \\
            &  & (512, 50, 50) $\times$ 3 &  &  &  &  \\ 
            
            \midrule 
            
            \multirow{3}{*}{Squeezing Block $f_{3}^{\text{sqz}}$} & \multirow{3}{*}{$\mathbf{C}_{3}$} & \multirow{3}{*}{(1, 12, 12, 12, 12)} & \multirow{3}{*}{$\mathbf{C}_{3}^{\text{sqz}}$} & \multirow{3}{*}{(128, 12, 12, 2, 2)} & \multirow{3}{*}{$\begin{pmatrix}\textsc{CP 4D conv}\\\textsc{Group Norm}\\\textsc{ReLU}\end{pmatrix} \times 3$} & \multirow{3}{*}{167K} \\
            & & & & & & \\
            & & & & & & \\
             
            \midrule 
            
            \multirow{3}{*}{Squeezing Block $f_{2}^{\text{sqz}}$} & \multirow{3}{*}{$\mathbf{C}_{2}$} & \multirow{3}{*}{(3, 25, 25, 25, 25)} & \multirow{3}{*}{$\mathbf{C}_{2}^{\text{sqz}}$} & \multirow{3}{*}{(128, 25, 25, 2, 2)} & \multirow{3}{*}{$\begin{pmatrix}\textsc{CP 4D conv}\\\textsc{Group Norm}\\\textsc{ReLU}\end{pmatrix} \times 3$} & \multirow{3}{*}{169K} \\
            & & & & & & \\
            & & & & & & \\
             
            \midrule 
            
            \multirow{3}{*}{Squeezing Block $f_{1}^{\text{sqz}}$} & \multirow{3}{*}{$\mathbf{C}_{1}$} & \multirow{3}{*}{(3, 50, 50, 50, 50)} & \multirow{3}{*}{$\mathbf{C}_{1}^{\text{sqz}}$} & \multirow{3}{*}{(128, 50, 50, 2, 2)} & \multirow{3}{*}{$\begin{pmatrix}\textsc{CP 4D conv}\\\textsc{Group Norm}\\\textsc{ReLU}\end{pmatrix} \times 3$} & \multirow{3}{*}{202K} \\
            & & & & & & \\
            & & & & & & \\
             
            \midrule 
            
            \multirow{6}{*}{Mixing Block $f_{2}^{\text{mix}}$} & & & \multirow{6}{*}{$\mathbf{C}_{2}^{\text{mix}}$} & \multirow{6}{*}{(128, 25, 25, 2, 2)} & \multirow{6}{*}{\shortstack{\textsc{Bilinear Interpolation}\\\textsc{Element-wise Addition}\\$\begin{pmatrix}\textsc{CP 4D conv}\\\textsc{Group Norm}\\\textsc{ReLU}\end{pmatrix} \times 3$}} & \multirow{6}{*}{886K} \\
            & $\mathbf{C}_{3}^{\text{sqz}}$ & (128, 12, 12, 2, 2) & & & & \\
            & & & & & & \\\cline{2-3}
            & & & & & & \\
            & $\mathbf{C}_{2}^{\text{sqz}}$ & (128, 25, 25, 2, 2) & & & & \\
            & & & & & & \\
             
            \midrule 
            
            \multirow{6}{*}{Mixing Block $f_{1}^{\text{mix}}$} & & & \multirow{6}{*}{$\mathbf{C}_{1}^{\text{mix}}$} & \multirow{6}{*}{(128, 50, 50, 2, 2)} & \multirow{6}{*}{\shortstack{\textsc{Bilinear Interpolation}\\\textsc{Element-wise Addition}\\$\begin{pmatrix}\textsc{CP 4D conv}\\\textsc{Group Norm}\\\textsc{ReLU}\end{pmatrix} \times 3$}} & \multirow{6}{*}{886K} \\
            & $\mathbf{C}_{2}^{\text{mix}}$ & (128, 25, 25, 2, 2) & & & & \\
            & & & & & & \\\cline{2-3}
            & & & & & & \\
            & $\mathbf{C}_{1}^{\text{sqz}}$ & (128, 50, 50, 2, 2) & & & & \\
            & & & & & & \\
             
            \midrule 
            
            Pooling Layer & $\mathbf{C}_{1}^{\text{mix}}$ & (128, 50, 50, 2, 2) & $\mathbf{Z}$ & (128, 50, 50) & \textsc{Average-pooling} & - \\
            
            \midrule
            
            \multirow{2}{*}{Decoder Layer} & \multirow{2}{*}{$\mathbf{Z}$} & \multirow{2}{*}{(128, 50, 50)} & \multirow{2}{*}{$\mathbf{\hat{M}}^{\mathrm{q}}$} & \multirow{2}{*}{(2, 400, 400)} & \textsc{Series of 2D Convs with} & \multirow{2}{*}{259K}\\
            & & & & & \textsc{Bilinear Interpolation} & \\

            \bottomrule
    \end{tabular}
    }
    \caption{\label{tab:vgg16_arch}Hypercorrelation Squeeze Networks (HSNet) with VGG16~\cite{simonyan2015vgg} backbone network. The reported number of parameters in VGG16 backbone network (14.7M) excludes those in fully-connected layers (unused in our model). The total number of `learnable' parameters amounts to 2.6M. The number of intermediate features extracted from backbone network amounts to 7, \ie, $L=7$. The \textsc{CP 4D conv} refers to the proposed center-pivot 4D convolution.}
    \end{center}
\end{table*}

\clearpage

\begin{table*}
    \begin{center}
    \scalebox{0.88}{
    \begin{tabular}{lcccccc}
            \toprule
            \textbf{Layer} & \multicolumn{2}{c}{\textbf{Input}} & \multicolumn{2}{c}{\textbf{Output}} & \textbf{Operation} & \textbf{\# params.} \\
            
            \midrule
            
            \multirow{6}{*}{ResNet50 Backbone} & \multirow{3}{*}{$I^{\mathrm{q}}$} & \multirow{3}{*}{(3, 400, 400)} & \multirow{3}{*}{$\{\mathbf{F}_{l}^{\mathrm{q}}\}_{l=1}^{13}$} & (2048, 13, 13) $\times$ 3 & \multirow{6}{*}{\textsc{Series of 2D Convs}} & \multirow{6}{*}{\shortstack{23.6M\\(frozen)}} \\
            
            & & & & (1024, 25, 25) $\times$ 6 & & \\
            & & & & (512, 50, 50) $\times$ 4 & & \\ \cline{2-5} \\[-2.0ex]
            
            & \multirow{3}{*}{$I^{\mathrm{s}}$} & \multirow{3}{*}{(3, 400, 400)} & \multirow{3}{*}{$\{\mathbf{F}_{l}^{\mathrm{s}}\}_{l=1}^{13}$} & (2048, 13, 13) $\times$ 3 &  &  \\
            & & & & (1024, 25, 25) $\times$ 6 & & \\
            & & & & (512, 50, 50) $\times$ 4 & & \\
            
            \midrule
            
            \multirow{4}{*}{Masking Layer} & \multirow{3}{*}{$\{\mathbf{F}_{l}^{\mathrm{s}}\}_{l=1}^{13}$} & (2048, 13, 13) $\times$ 3 & \multirow{4}{*}{$\{\hat{\mathbf{F}}_{l}^{\mathrm{s}}\}_{l=1}^{13}$} & \multirow{4}{*}{\shortstack{(2048, 13, 13) $\times$ 3\\(1024, 25, 25) $\times$ 6\\(512, 50, 50) $\times$ 4}} & \multirow{4}{*}{\shortstack{\textsc{Bilinear Interpolation}\\\textsc{Hadamard Product}}} & \multirow{4}{*}{-} \\
            
            &  & (1024, 25, 25) $\times$ 6 &  &  &  &  \\
            &  & (512, 50, 50) $\times$ 4 &  &  &  &  \\ \cline{2-3} \\[-2.0ex]
            & $\mathbf{M}^{\mathrm{s}}$ & (1, 400, 400) &  &  &  &  \\
            
            \midrule
            
            \multirow{6}{*}{Correlation Layer} & \multirow{3}{*}{$\{\mathbf{F}_{l}^{\mathrm{q}}\}_{l=1}^{13}$} & (2048, 13, 13) $\times$ 3 & \multirow{6}{*}{$\{\mathbf{C}_{p}\}_{p=1}^{3}$} & \multirow{6}{*}{\shortstack{(3, 13, 13, 13, 13)\\(6, 25, 25, 25, 25)\\(4, 50, 50, 50, 50)}} & \multirow{6}{*}{\textsc{Cosine Similarity}} & \multirow{6}{*}{-} \\
            
            &  & (1024, 25, 25) $\times$ 6 &  &  &  &  \\
            &  & (512, 50, 50) $\times$ 4 &  &  &  &  \\ \cline{2-3} \\[-2.0ex]
            & \multirow{3}{*}{$\{\hat{\mathbf{F}}_{l}^{\mathrm{s}}\}_{l=1}^{13}$} & (2048, 13, 13) $\times$ 3 &  &  &  &  \\
            &  & (1024, 25, 25) $\times$ 6 &  &  &  &  \\
            &  & (512, 50, 50) $\times$ 4 &  &  &  &  \\ 
            
            \midrule 
            
            \multirow{3}{*}{Squeezing Block $f_{3}^{\text{sqz}}$} & \multirow{3}{*}{$\mathbf{C}_{3}$} & \multirow{3}{*}{(3, 13, 13, 13, 13)} & \multirow{3}{*}{$\mathbf{C}_{3}^{\text{sqz}}$} & \multirow{3}{*}{(128, 13, 13, 2, 2)} & \multirow{3}{*}{$\begin{pmatrix}\textsc{CP 4D conv}\\\textsc{Group Norm}\\\textsc{ReLU}\end{pmatrix} \times 3$} & \multirow{3}{*}{168K} \\
            & & & & & & \\
            & & & & & & \\
             
            \midrule 
            
            \multirow{3}{*}{Squeezing Block $f_{2}^{\text{sqz}}$} & \multirow{3}{*}{$\mathbf{C}_{2}$} & \multirow{3}{*}{(6, 25, 25, 25, 25)} & \multirow{3}{*}{$\mathbf{C}_{2}^{\text{sqz}}$} & \multirow{3}{*}{(128, 25, 25, 2, 2)} & \multirow{3}{*}{$\begin{pmatrix}\textsc{CP 4D conv}\\\textsc{Group Norm}\\\textsc{ReLU}\end{pmatrix} \times 3$} & \multirow{3}{*}{172K} \\
            & & & & & & \\
            & & & & & & \\
             
            \midrule 
            
            \multirow{3}{*}{Squeezing Block $f_{1}^{\text{sqz}}$} & \multirow{3}{*}{$\mathbf{C}_{1}$} & \multirow{3}{*}{(4, 50, 50, 50, 50)} & \multirow{3}{*}{$\mathbf{C}_{1}^{\text{sqz}}$} & \multirow{3}{*}{(128, 50, 50, 2, 2)} & \multirow{3}{*}{$\begin{pmatrix}\textsc{CP 4D conv}\\\textsc{Group Norm}\\\textsc{ReLU}\end{pmatrix} \times 3$} & \multirow{3}{*}{203K} \\
            & & & & & & \\
            & & & & & & \\
             
            \midrule 
            
            \multirow{6}{*}{Mixing Block $f_{2}^{\text{mix}}$} & & & \multirow{6}{*}{$\mathbf{C}_{2}^{\text{mix}}$} & \multirow{6}{*}{(128, 25, 25, 2, 2)} & \multirow{6}{*}{\shortstack{\textsc{Bilinear Interpolation}\\\textsc{Element-wise Addition}\\$\begin{pmatrix}\textsc{CP 4D conv}\\\textsc{Group Norm}\\\textsc{ReLU}\end{pmatrix} \times 3$}} & \multirow{6}{*}{886K} \\
            & $\mathbf{C}_{3}^{\text{sqz}}$ & (128, 13, 13, 2, 2) & & & & \\
            & & & & & & \\\cline{2-3}
            & & & & & & \\
            & $\mathbf{C}_{2}^{\text{sqz}}$ & (128, 25, 25, 2, 2) & & & & \\
            & & & & & & \\
             
            \midrule 
            
            \multirow{6}{*}{Mixing Block $f_{1}^{\text{mix}}$} & & & \multirow{6}{*}{$\mathbf{C}_{1}^{\text{mix}}$} & \multirow{6}{*}{(128, 50, 50, 2, 2)} & \multirow{6}{*}{\shortstack{\textsc{Bilinear Interpolation}\\\textsc{Element-wise Addition}\\$\begin{pmatrix}\textsc{CP 4D conv}\\\textsc{Group Norm}\\\textsc{ReLU}\end{pmatrix} \times 3$}} & \multirow{6}{*}{886K} \\
            & $\mathbf{C}_{2}^{\text{mix}}$ & (128, 25, 25, 2, 2) & & & & \\
            & & & & & & \\\cline{2-3}
            & & & & & & \\
            & $\mathbf{C}_{1}^{\text{sqz}}$ & (128, 50, 50, 2, 2) & & & & \\
            & & & & & & \\
             
            \midrule 
            
            Pooling Layer & $\mathbf{C}_{1}^{\text{mix}}$ & (128, 50, 50, 2, 2) & $\mathbf{Z}$ & (128, 50, 50) & \textsc{Average-pooling} & - \\
            
            \midrule
            
            \multirow{2}{*}{Decoder Layer} & \multirow{2}{*}{$\mathbf{Z}$} & \multirow{2}{*}{(128, 50, 50)} & \multirow{2}{*}{$\mathbf{\hat{M}}^{\mathrm{q}}$} & \multirow{2}{*}{(2, 400, 400)} & \textsc{Series of 2D Convs with} & \multirow{2}{*}{259K}\\
            & & & & & \textsc{Bilinear Interpolation} & \\

            \bottomrule
    \end{tabular}
    }
    \caption{\label{tab:res50_arch}Hypercorrelation Squeeze Networks (HSNet) with ResNet50~\cite{he2016deep} backbone network. The reported number of parameters in ResNet50 backbone network (23.6M) excludes those in fully-connected layers (unused in our model). The total number of `learnable' parameters amounts to 2.6M. The number of intermediate features extracted from backbone network amounts to 13, \ie, $L=13$.}
    \end{center}
\end{table*}

\clearpage

\begin{table*}
    \begin{center}
    \scalebox{0.88}{
    \begin{tabular}{lcccccc}
            \toprule
            \textbf{Layer} & \multicolumn{2}{c}{\textbf{Input}} & \multicolumn{2}{c}{\textbf{Output}} & \textbf{Operation} & \textbf{\# params.} \\
            
            \midrule
            
            \multirow{6}{*}{ResNet101 Backbone} & \multirow{3}{*}{$I^{\mathrm{q}}$} & \multirow{3}{*}{(3, 400, 400)} & \multirow{3}{*}{$\{\mathbf{F}_{l}^{\mathrm{q}}\}_{l=1}^{30}$} & (2048, 13, 13) $\times$ 3 & \multirow{6}{*}{\textsc{Series of 2D Convs}} & \multirow{6}{*}{\shortstack{42.6M\\(frozen)}} \\
            
            & & & & (1024, 25, 25) $\times$ 23 & & \\
            & & & & (512, 50, 50) $\times$ 4 & & \\ \cline{2-5} \\[-2.0ex]
            
            & \multirow{3}{*}{$I^{\mathrm{s}}$} & \multirow{3}{*}{(3, 400, 400)} & \multirow{3}{*}{$\{\mathbf{F}_{l}^{\mathrm{s}}\}_{l=1}^{30}$} & (2048, 13, 13) $\times$ 3 &  &  \\
            & & & & (1024, 25, 25) $\times$ 23 & & \\
            & & & & (512, 50, 50) $\times$ 4 & & \\
            
            \midrule
            
            \multirow{4}{*}{Masking Layer} & \multirow{3}{*}{$\{\mathbf{F}_{l}^{\mathrm{s}}\}_{l=1}^{30}$} & (2048, 13, 13) $\times$ 3 & \multirow{4}{*}{$\{\hat{\mathbf{F}}_{l}^{\mathrm{s}}\}_{l=1}^{30}$} & \multirow{4}{*}{\shortstack{(2048, 13, 13) $\times$ 3\\(1024, 25, 25) $\times$ 23\\(512, 50, 50) $\times$ 4}} & \multirow{4}{*}{\shortstack{\textsc{Bilinear Interpolation}\\\textsc{Hadamard Product}}} & \multirow{4}{*}{-} \\
            
            &  & (1024, 25, 25) $\times$ 23 &  &  &  &  \\
            &  & (512, 50, 50) $\times$ 4 &  &  &  &  \\ \cline{2-3} \\[-2.0ex]
            & $\mathbf{M}^{\mathrm{s}}$ & (1, 400, 400) &  &  &  &  \\
            
            \midrule
            
            \multirow{6}{*}{Correlation Layer} & \multirow{3}{*}{$\{\mathbf{F}_{l}^{\mathrm{q}}\}_{l=1}^{30}$} & (2048, 13, 13) $\times$ 3 & \multirow{6}{*}{$\{\mathbf{C}_{p}\}_{p=1}^{3}$} & \multirow{6}{*}{\shortstack{(3, 13, 13, 13, 13)\\(23, 25, 25, 25, 25)\\(4, 50, 50, 50, 50)}} & \multirow{6}{*}{\textsc{Cosine Similarity}} & \multirow{6}{*}{-} \\
            
            &  & (1024, 25, 25) $\times$ 23 &  &  &  &  \\
            &  & (512, 50, 50) $\times$ 4 &  &  &  &  \\ \cline{2-3} \\[-2.0ex]
            & \multirow{3}{*}{$\{\hat{\mathbf{F}}_{l}^{\mathrm{s}}\}_{l=1}^{30}$} & (2048, 13, 13) $\times$ 3 &  &  &  &  \\
            &  & (1024, 25, 25) $\times$ 23 &  &  &  &  \\
            &  & (512, 50, 50) $\times$ 4 &  &  &  &  \\ 
            
            \midrule 
            
            \multirow{3}{*}{Squeezing Block $f_{3}^{\text{sqz}}$} & \multirow{3}{*}{$\mathbf{C}_{3}$} & \multirow{3}{*}{(3, 13, 13, 13, 13)} & \multirow{3}{*}{$\mathbf{C}_{3}^{\text{sqz}}$} & \multirow{3}{*}{(128, 13, 13, 2, 2)} & \multirow{3}{*}{$\begin{pmatrix}\textsc{CP 4D conv}\\\textsc{Group Norm}\\\textsc{ReLU}\end{pmatrix} \times 3$} & \multirow{3}{*}{168K} \\
            & & & & & & \\
            & & & & & & \\
             
            \midrule 
            
            \multirow{3}{*}{Squeezing Block $f_{2}^{\text{sqz}}$} & \multirow{3}{*}{$\mathbf{C}_{2}$} & \multirow{3}{*}{(23, 25, 25, 25, 25)} & \multirow{3}{*}{$\mathbf{C}_{2}^{\text{sqz}}$} & \multirow{3}{*}{(128, 25, 25, 2, 2)} & \multirow{3}{*}{$\begin{pmatrix}\textsc{CP 4D conv}\\\textsc{Group Norm}\\\textsc{ReLU}\end{pmatrix} \times 3$} & \multirow{3}{*}{185K} \\
            & & & & & & \\
            & & & & & & \\
             
            \midrule 
            
            \multirow{3}{*}{Squeezing Block $f_{1}^{\text{sqz}}$} & \multirow{3}{*}{$\mathbf{C}_{1}$} & \multirow{3}{*}{(4, 50, 50, 50, 50)} & \multirow{3}{*}{$\mathbf{C}_{1}^{\text{sqz}}$} & \multirow{3}{*}{(128, 50, 50, 2, 2)} & \multirow{3}{*}{$\begin{pmatrix}\textsc{CP 4D conv}\\\textsc{Group Norm}\\\textsc{ReLU}\end{pmatrix} \times 3$} & \multirow{3}{*}{203K} \\
            & & & & & & \\
            & & & & & & \\
             
            \midrule 
            
            \multirow{6}{*}{Mixing Block $f_{2}^{\text{mix}}$} & & & \multirow{6}{*}{$\mathbf{C}_{2}^{\text{mix}}$} & \multirow{6}{*}{(128, 25, 25, 2, 2)} & \multirow{6}{*}{\shortstack{\textsc{Bilinear Interpolation}\\\textsc{Element-wise Addition}\\$\begin{pmatrix}\textsc{CP 4D conv}\\\textsc{Group Norm}\\\textsc{ReLU}\end{pmatrix} \times 3$}} & \multirow{6}{*}{886K} \\
            & $\mathbf{C}_{3}^{\text{sqz}}$ & (128, 13, 13, 2, 2) & & & & \\
            & & & & & & \\\cline{2-3}
            & & & & & & \\
            & $\mathbf{C}_{2}^{\text{sqz}}$ & (128, 25, 25, 2, 2) & & & & \\
            & & & & & & \\
             
            \midrule 
            
            \multirow{6}{*}{Mixing Block $f_{1}^{\text{mix}}$} & & & \multirow{6}{*}{$\mathbf{C}_{1}^{\text{mix}}$} & \multirow{6}{*}{(128, 50, 50, 2, 2)} & \multirow{6}{*}{\shortstack{\textsc{Bilinear Interpolation}\\\textsc{Element-wise Addition}\\$\begin{pmatrix}\textsc{CP 4D conv}\\\textsc{Group Norm}\\\textsc{ReLU}\end{pmatrix} \times 3$}} & \multirow{6}{*}{886K} \\
            & $\mathbf{C}_{2}^{\text{mix}}$ & (128, 25, 25, 2, 2) & & & & \\
            & & & & & & \\\cline{2-3}
            & & & & & & \\
            & $\mathbf{C}_{1}^{\text{sqz}}$ & (128, 50, 50, 2, 2) & & & & \\
            & & & & & & \\
             
            \midrule 
            
            Pooling Layer & $\mathbf{C}_{1}^{\text{mix}}$ & (128, 50, 50, 2, 2) & $\mathbf{Z}$ & (128, 50, 50) & \textsc{Average-pooling} & - \\
            
            \midrule
            
            \multirow{2}{*}{Decoder Layer} & \multirow{2}{*}{$\mathbf{Z}$} & \multirow{2}{*}{(128, 50, 50)} & \multirow{2}{*}{$\mathbf{\hat{M}}^{\mathrm{q}}$} & \multirow{2}{*}{(2, 400, 400)} & \textsc{Series of 2D Convs with} & \multirow{2}{*}{259K}\\
            & & & & & \textsc{Bilinear Interpolation} & \\
            
            \bottomrule
    \end{tabular}
    }
    \caption{\label{tab:res101_arch}Hypercorrelation Squeeze Networks (HSNet) with ResNet101~\cite{he2016deep} backbone network. The reported number of parameters in ResNet101 backbone network (42.6M) excludes those in fully-connected layers (unused in our model). The total number of `learnable' parameters amounts to 2.6M. The number of intermediate features extracted from backbone network amounts to 30, \ie, $L=30$.}
    \end{center}
\end{table*}

\end{appendices}

\clearpage

\end{document}